\documentclass[manuscript]{acmart}

\setcopyright{none}
\renewcommand\footnotetextcopyrightpermission[1]{} 
\usepackage{amsmath}
\usepackage{multirow}
\usepackage{tabularx}
\usepackage{url}

\usepackage{hyperref}

\usepackage{longtable}

\usepackage[justification=centering]{caption}

\author{Tommi Gr\"{o}ndahl}
\author{N. Asokan}

\title{Text Analysis in Adversarial Settings: Does Deception Leave a Stylistic Trace?}

\begin{document}

\begin{abstract}

Textual deception constitutes a major problem for online security. Many studies have argued that deceptiveness leaves traces in writing style, which could be detected using text classification techniques. By conducting an extensive literature review of existing empirical work, we demonstrate that while certain linguistic features have been indicative of deception in certain corpora, they fail to generalize across divergent semantic domains. We suggest that deceptiveness as such leaves no \textit{content-invariant stylistic trace}, and textual similarity measures provide superior means of classifying texts as potentially deceptive. Additionally, we discuss forms of deception beyond semantic content, focusing on hiding author identity by \textit{writing style obfuscation}. Surveying the literature on both author identification and obfuscation techniques, we conclude that current style transformation methods fail to achieve reliable obfuscation while simultaneously ensuring semantic faithfulness to the original text. We propose that future work in style transformation should pay particular attention to disallowing semantically drastic changes.

\end{abstract}

\maketitle

\section{Introduction}
\label{sec:intro}

Deception is rampant in online text, and its detection constitutes a major challenge at the crossroads of natural language processing (NLP) and information security research. Multiple studies have contended that leading machine learning techniques are able to extract features that can distinguish between deceptive and normal text. In order for such features to truly reflect deceptiveness instead of domain-specific lexical content, the features discovered should generalize across domains.
Variants of deception also extend beyond textual content. In particular, metalinguistic information can be obfuscated to deceive a classifier while retaining the original content. Of such endeavours, the most prominently discussed has been \textit{adversarial stylometry} \cite{Brennan:Greenstadt2009, Brennanetal2011}, consisting of techniques that attempt to provide author anonymity by defeating identification or profiling. In this survey we review prior research on textual deception and its detection, focusing on deceptive content in Section \ref{sec:deception} and adversarial stylometry in Section \ref{sec:author-identification}.

Modern NLP techniques offer a large variety of methods for classifying texts based on the distribution of \emph{linguistic information}: features that are detectable from \textit{text alone}, without extra-linguistic knowledge concerning author behavior, metadata etc. Depending on the task, target categories can be delineated by semantic content, grammar, or any combination of these. Identifying or profiling authors based on writing style comprises the field of \textit{stylometry}.
As a scientific endeavour it dates back at least to the 19th century \cite{Lutoslawski1898, Mendenhall1887}, and was formulated as a computational task in the 1960s \cite{Mosteller:Wallace1964a, Stamatatos2009}. In contemporary work, the traditional focus on literary documents has largely been overshadowed by the increased use of online datasets, such as blog posts \cite{Narayanetal2012}, e-mails \cite{DeVeletal2001, Crabb2014}, forum discussions \cite{Zhengetal2006}, SMS messages \cite{Rageletal2013}, and tweets \cite{Castro:Lindauer2013}.
Neal et al. \cite{Nealetal2017} comprehensively survey the state-of-the-art in stylometry.

Stylometry uses linguistic information to extract a \textit{non-linguistic} property of the author of a text, such as identity, gender or age. Within NLP, it thus belongs to the field of \textit{metaknowledge extraction} \cite{Daelemans2013}, which relies on linguistic information systematically correlating with the relevant property under investigation, despite that property itself not being linguistic. The conjecture that author identity can be reliably inferred from his/her stylistic ``fingerprint'' is known as the \textit{Human  Stylome Hypothesis} (HSH) \cite{vanHalterenetal2005}.

\hypertarget{link:stylistic-trace}{Motivated} by the HSH, we can also formulate an analogous question about any other property of a text: does the property leave a \textit{linguistic trace}, and if so, to what extent does it leave a content-independent \textit{stylistic trace} that could be recognized across semantically distant texts?
In this paper we discuss this question with respect to a class of properties that fall under the umbrella term of \textit{deception}. We investigate the issue both from the perspective of detecting deception in text, and from the \textit{adversarial} perspective of creating deceptive data that can evade classification. Our focus is on information security applications in particular. Terminologically, we call non-deceptive text ``normal''.

If reliable linguistic cues of deception existed, they could be used to detect security breaches such as fake reviews \cite{Yoo:Gretzel2009, Ottetal2011, Ottetal2013, Lietal2014, Xuetal2015}, troll-messages \cite{Cambriaetal2010, Seahetal2015, Mihaylov:Nakov2016}, or even fake news \cite{Perez-Rosasetal2018, Oshikawaetal2018}. A number of prior studies have attempted to demonstrate the potential of stylometry for deception detection, and to find the major linguistic determinants of deceptive text. We review and discuss this research in Section \ref{sec:deception}. A common assumption behind all these studies is that deception leaves a stylistic trace comparable to an author's ``stylome''. If true, this would allow detecting deceptiveness \textit{from the text alone}, without recourse to extra-linguistic information. If, on the other hand, deception leaves no major stylistic trace, its reliable detection would require linguistic analyses to be augmented with other techniques. Based on the survey, we argue for the latter position, and suggest alternative methods based on \textit{content-comparison} that provide more promising approaches for this task (Section \ref{sec:deception-future-prospects}).

In Section \ref{sec:author-identification} we turn to adversarial stylometry.
From a security perspective, it simultaneously functions as an \textit{attack} against authorship classification, and as a \textit{defence} against non-consensual deanonymization or profiling. The latter scenario has been called the \textit{deanonymization attack} \cite{Narayanetal2012}, and its feasibility is conditional on the HSH. Therefore a major question is whether current author identification techniques pose a realistic privacy threat. Based on a review of state-of-the-art stylometry research in Section \ref{sec:author-identification-general}, we argue that while the HSH has not always been validated, the deanonymization attack constitutes a genuine privacy concern especially when the candidate authors are few in number. In Section \ref{sec:deanonymization} we discuss the attack scenario in more detail.

Methods for style transformation can be divided into manual, computer-assisted and automatic techniques. For \textit{ordinary} users, only the last would constitute a practically effective mitigation against the deanonymization attack. Manual obfuscation is difficult and time-consuming, and requires a good grasp of linguistic subtleties, which makes the task unsuitable for users lacking extra time and resources. An additional difference can be made between \textit{obfuscation} and \textit{imitation}, where the latter targets a particular style instead of simply avoiding detection.
Section \ref{sec:adversarial_stylometry} reviews existing work on style obfuscation and imitation techniques. We argue that while some methods show potential in principle, all face serious problems with balancing between obfuscation and maintaining semantic content.

Given that style obfuscation and imitation constitute types of deception, we then return to the original question of whether deception leaves a stylistic trace, and apply it to this special case. Even if obfuscation was successful, the property of \textit{being obfuscated} could itself be stylometrically traceable.
Problematically, our review in Section \ref{sec:automatic-obfuscation} demonstrates that this question has typically not been tested. Studies attempting such ``fingerprinting'' of the obfuscation method have succeeded \cite{Caliskan:Greenstadt2012, Dayetal2016}, but have only experimented on a small subset of possible methods. As different techniques require different recognition methods, a general detector of style obfuscation is likely difficult to attain.


In summary, this survey addresses three major \hypertarget{link:intro-questions}{questions}:

\begin{itemize}
\item[Q1] Does deception leave a content-independent stylistic trace?
\item[Q2] Is the deanonymization attack a realistic privacy concern?
\item[Q3] Can the deanonymization attack be mitigated with automatic style obfuscation?
\end{itemize}

Q1 provides the common theme of the survey. Section \ref{sec:deception} discusses the linguistic detection of \textit{deceptive content}, with a particular focus on online text. Section \ref{sec:author-identification} then moves on to the topic of adversarial stylometry, i.e. mitigating the deanonymization attack (Q2) via style transformation (Q3).

We summarize our findings and suggestions below.

\begin{itemize}
\item There is no evidence that deception leaves a \textit{content-invariant} stylistic trace.
Instead, detection should involve the \textit{comparison of semantic content} across texts.
\item While the validity of the HSH is uncertain, the deanonymization attack is a realistic privacy concern.
\item As of yet, automatic style transformation techniques do not secure semantic faithfulness.
\end{itemize}

\section{Deception detection via text analysis}
\label{sec:deception}

%

In this section we review the research on textual deception detection, and discuss the linguistic features associated with writing intended to deceive the reader. 
Multiple studies have indicated that at least non-expert human accuracy in detecting textual deception is approximately on a chance level, or even worse \cite{Ekman:Osullivan1991, Newmanetal2003, Bond:Paulo2006, Enosetal2007}. As Fitzpatrick et al. \cite{Fitzpatricketal2015} note, this makes deception detection a somewhat exceptional topic for NLP, since human performance in most other text classification problems tends to be more accurate than computational solutions. In contrast, automated classification of deceptive text should increase not only the efficiency but also the accuracy of human performance.
However, we argue that the divergence of features deemed relevant by different studies indicates that classification has been too content-specific to generalize across semantic domains. Relevant features tend to be lexical correlates of deceptive text in particular corpora rather than general ``deception markers'' as such.

Most research in deception detection has concerned face-to-face discussion \cite{Ekman1985, Ekman:Osullivan1991, DePauloetal2003, Enosetal2007, Fitzpatrick:Bachenko2009, Fitzpatricketal2015}. As Crabb \cite{Crabb2014} notes, such results do not always directly apply to communication via electronic devices, which are the most relevant for information security concerns. In particular, physiological data is unavailable to the receiver in text-based communication.
We limit our discussion to deceptive communication in written English.
Following DePaulo et al. \cite{DePauloetal2003}, we dissociate deceptiveness as a \textit{communicative intention} from falsity as a semantic property.\footnote{\label{fn:deception_literal} Literal truths with a deceptive intention include cases where the speaker believes the hearer to \textit{infer} a falsity from a literal truth. If Bob asks: ``Where is Jim?'', and Alice answers: ``I saw him in the cafeteria'', in normal circumstances Alice assumes Bob to infer that Jim may still be there. Hence, if Alice believes Jim not to be there anymore (e.g. she also saw him leave the cafeteria), she is deceiving Bob by telling a literal truth. Assumed inferences can be understood as belonging to communicated content as \textit{implicatures} \cite{Sperber:Wilson95}.}
Utilizing a famous formulation by Paul Grice, \emph{communication} can roughly be characterized as behavior with the deliberate goal of causing certain thoughts in the (intended) receiver \cite{Grice89, Sperber:Wilson95}. Deception thus constitutes a specific type of communication, where the speaker intends the hearer to form thoughts which the speaker believes to be false. The notion of deception as an author intention is also shared by Buller and Burgoon's \emph{Interpersonal Deception Theory} \cite{Buller:Burgoon1996a}.
For our purposes, we can use the following general characterization:

\begin{itemize}
\item[] \textbf{Deception}
\item[] A deceives B if for some proposition P:
\begin{itemize}
\item[] A believes that P is false
\item[] A attempts to make B believe that P is true
\end{itemize}
\end{itemize}

The deceptiveness of a communicative act makes no restrictions on the nature of the proposition P. In particular, P might not belong to the semantic content of the expression E.
We can thus separate between \textit{explicit} and \emph{implicit deception} as follows:

\begin{itemize}
\item[] \textbf{Explicit deception}
\item[] A explicitly deceives B if for some proposition P:
\begin{itemize}
\item[] A believes that P is false
\item[] A attempts to make B believe that P is true by uttering an expression E
\item[] The semantic content of E contains P
\end{itemize}
\end{itemize}

\begin{itemize}
\item[] \textbf{Implicit deception}
\item[] A implicitly deceives B if for some proposition P:
\begin{itemize}
\item[] A believes that P is false
\item[] A attempts to make B believe that P is true by uttering an expression E
\item[] The semantic content of E does not contain P
\item[] A assumes B to infer P from the explicit content of E and context information that A assumes B to know
\end{itemize}
\end{itemize}

In both cases, B infers P from A's utterance E. In explicit deception, P can be found directly from E itself without consulting other assumptions or beliefs within the discourse. In implicit deception, further inferences are needed to come to the conclusion P.
As an example, consider fake online reviews, which Section \ref{sec:fake-reviews} will discuss in detail. Some fake reviews contain explicit falsities: if a TripAdvisor user claims to have been in a hotel and (dis)liked it, this is explicitly deceptive if he actually has not visited the hotel. However, suppose the reviewer merely makes general claims about the hotel (``This hotel is excellent/horrible!'' etc.). Here, the deception concerns the reviewer's first-hand experience, which he lacks, and is independent of the reviewer's actual beliefs of his review's correctness. Therefore, it can be treated as a variant of implicit deception.

We further divide different types of deception reviewed in Sections \ref{sec:deception-general}--\ref{sec:deception-security} to the following three groups, the first being explicit and the two latter implicit:

\begin{itemize}
\item[] \textbf{Deception of literal content:} the semantic content of the text is deceptive
\item[] \textbf{Deception of authority:} the deceiver implies having authority concerning the issue, which he lacks
\item[] \textbf{Deception of intention:} the deceiver has an ulterior deceptive motive for writing the message
\end{itemize}

While not meant to be exhaustive, this taxonomy is useful in accounting for disparities between different studies.
Most studies reviewed concern deception of literal content. However, as argued above, fake reviews can exhibit deception of \textit{authority} instead.
Deception of intention is exemplified by \textit{trolling}, where the author writes something to advance a particular view or to harass another person. Here, deception does not necessarily concern the literal content (which may sometimes be sincerely believed by the troll), but instead the ulterior motive behind the message.

Ultimately, the issue at hand is whether it would be possible to develop a ``textual lie detector'' that takes a text as an input and outputs a classification label that reliably tracks the real-world property of deceptiveness. The main problem for such a goal is that even if deceptive texts differ from non-deceptive texts in particular corpora, the features may not generalize across different text types. Deception could leave \textit{some} stylistic cues in e.g. in online discussions, fake news, fake reviews, or scientific papers; but in order for the hypothetical ``lie detector'' to work, these cues should be sufficiently \textit{similar} across them all.
To evaluate whether existing methods are applicable for such general deception detection, we need to compare empirical results from different studies and see if common patterns emerge.

Table \ref{tab:deception-summary} shows the linguistic properties that appear three or more times in the studies reviewed in Sections \ref{sec:deception-general}-\ref{sec:deception-security}. Based on these, we formulate the following hypotheses:

\begin{table}
  \begin{center}
    \begin{tabular}{|c|c|} \hline
    \textbf{Studies} & \textbf{Cue} \\ \hline
    
    \cite{Burgoonetal2003, Zhouetal2004a, Zhouetal2004b, Newmanetal2003, Keila:Skillicorn05, Hancock:Markowitz2014, Ottetal2011, Ottetal2013, Lietal2014}
    & High emotional load \\ \hline
    
    \cite{Burgoonetal2003, Zhouetal2004a, Louwerseetal2010, Hancock:Markowitz2014, Ottetal2011, Ottetal2013, Xuetal2015}
    & Generality / abstractness / lack of specificity \\ \hline
    
    \cite{Leeetal2009, Yoo:Gretzel2009, Louwerseetal2010, Ottetal2011, Lietal2014}
    & High use of first-person pronouns \\ \hline
    
    \cite{Newmanetal2003, Keila:Skillicorn05, Hancocketal2008, Mihalcea:Strappava2009} 
    & Low use of first-person pronouns \\ \hline
    
    \cite{Zhouetal2003b, Ottetal2011, Ottetal2013, Crabb2014} 
    & High use of verbs \\ \hline
    
    \cite{Leeetal2009, Mihalcea:Strappava2009, Hancock:Markowitz2014}
    & High use of certainty-related words \\ \hline
    
    \end{tabular}
    \caption{The most common linguistic cues to deception from all studies reviewed in Sections \ref{sec:deception-general}-\ref{sec:deception-security}}
    \label{tab:deception-summary}
  \end{center}
\end{table}

\begin{itemize}
\item[\textbf{H1:}] deceptive text is emotionally laden
\item[\textbf{H2:}] deceptive text contains certainty-related terminology
\item[\textbf{H3:}] deceptive text lacks in detail
\item[\textbf{H4:}] deceptive text lacks a first-person narrative
\end{itemize}

Table \ref{tab:deception-summary} contains two contradictory properties: high and low use of first person pronouns. We choose the low use hypothesis as the default (H4), since it would be predicted by the lack of first-hand experience of the situation. This empirical divergence is indicative of the context-dependency of suggested deception cues.
Nevertheless, H1--H4 are intuitively understandable and fit well within standard psychological models of deception \cite{Buller:Burgoon1996a}. H1 can be explained either by the stress caused by lying \cite{Ekman:Friesen68}, or from attempted emotional persuasion of the audience. Experimental results reviewed in Section \ref{sec:fake-reviews} point to the latter \cite{Ottetal2011, Ottetal2013}. H2 also likely results from the persuasive purpose of deception. H3 and H4 are motivated by the fact that deceivers often have no first-hand experience of the situation they are describing.

H1--H4 can thus be argued to follow from two basic tendencies present in deception: \textit{attempted persuasion} and \textit{lack of first-hand knowledge}. Interestingly, these can sometimes motivate the deceiver to behave in \textit{opposite} ways, which may partly explain our seemingly inconsistent finding concerning first-person pronoun usage. Increased use of the first-person pronoun indicates a personal narrative and hence emphasizes the notion of the author having actually experienced the situation under discussion. It can therefore be used in an attempt to increase the credibility of the text. On the other hand, the lack of first-hand experience makes it more difficult for deceivers to credibly describe something they do not know in detail, and hence can motivate them to stick to a more general, third-person narrative.

Section \ref{sec:deception-general} reviews the literature on deception detection from a general perspective not specific to information security concerns. Section \ref{sec:deception-security} focuses specifically on deception in online text, discussing fake reviews (\ref{sec:fake-reviews}) and troll comments (\ref{sec:trolls}). We summarize our analyses and give recommendations for future research in Section \ref{sec:deception-future-prospects}.

\subsection{General deception detection}
\label{sec:deception-general}

In this section we focus on deception detection outside of online text datasets. We further distinguish between \textit{experimentally elicited} deception and natural or \textit{non-elicited} deception, devoting Section \ref{sec:deception_elicited} to the former and Section \ref{sec:deception-non-elicited} to the latter.

\subsubsection{Experimentally elicited deception}
\label{sec:deception_elicited}

In this section we present results from experimental research on elicited deceptive text. By \textit{elicited} we mean that the texts were produced at the command of a test instructor, and that their deceptiveness or truthfulness was explicitly requested.

Burgoon et al. \cite{Burgoonetal2003} formulated eight hypotheses concerning deception:

\begin{quote}
``deceptive senders display higher
\begin{itemize}
\item[(a)] quantity,
\item[(b)] non-immediacy,
\item[(c)] expressiveness,
\item[(d)] informality, and
\item[(e)] affect;
\end{itemize}
and less
\begin{itemize}
\item[(f)] complexity,
\item[(g)] diversity, and
\item[(h)] specificity of language'' \cite{Burgoonetal2003}\footnote{\emph{Quantity} means the amount of text produced, \emph{non-immediacy} refers to the lack of directness and intensity between the author and receiver, \emph{expressiveness} is the amount of descriptive material in the text (e.g. adjectives and adverbs), \emph{informality} is indicated e.g. by the amount of typos, \emph{affect} refers to the emotional load of the text, \emph{complexity} is measured by readability indices \cite{Smith:Senter67}, \emph{diversity} is the type-token ratio among words, and \emph{specificity} denotes the level of detail in the text.}
\end{itemize}
\end{quote}

They based these hypotheses on two experiments, where truthful and deceptive text was gathered from participants. The messages were obtained via e-mail in the first experiment, and via either face-to-face communication or text chat in the second experiment. However, the results of these experiments were contradictory, as deceivers used longer but less complex messages in the first test, and shorter but more complex messages in the second (although the results from the second test were not statistically significant). The hypotheses (a--h) reflect the results of the first experiment. In relation to H1--H4, (e) indicates \textit{emotional load} (H1) and (f--h) fall into the broader category of \textit{lacking detail} (H3). However, the latter is partly at odds with (c), which indicates that deceivers also use higher amounts of descriptive words. This effect could arise from the attempted persuasion involved in deception.

Based on nine linguistic properties similar to those suggested by Burgoon et al. \cite{Burgoonetal2003} (each composed of many features, 27 altogether), Zhou et al. \cite{Zhouetal2004a} classified experimentally elicited texts as deceptive or truthful. All features were relevant with the exception of specificity. In a subsequent study, Zhou et al. \cite{Zhouetal2004b} report 22 linguistic features as indicative of deception (see Table \ref{tab:deception_experimental}). Using these features, they compared four machine learning methods in the classification task: discriminant analysis, decision trees, neural networks and logistic regression. The methods fared roughly equally well, and at best achieved an accuracy of ca. $80\%$.

Newman et al.'s \cite{Newmanetal2003} results on classifying experimentally elicited deceptive and truthful texts indicated that deception was characterized by the reduced use of first- and third-person pronouns and exclusive words (e.g. \textit{but}, \textit{except}), along with the increased use of negative emotion words (e.g. \textit{hate}, \textit{anger}) and motion words (e.g. \textit{walk}, \textit{go}). These findings are partly in line with more general observations concerning deception, but not fully. While emotional load (H1) and reduced first-person pronoun use (H4) are expected , it is unclear why deception should correlate with the reduction of \textit{both} first- and third-person pronouns, a decrease in exclusive words, or an increase in motion words. Typically, first- and third-person pronoun usage can be seen as \textit{complementary} ways to talk about a situation, the first-person indicating a personal narrative and the third-person an impersonal one. It is therefore unclear what properties the reduction in \textit{both} would coincide with. Also, exclusion words are likely too abstract to be closely related to particular communicative functions, and hence it would be surprising if their prevalence in deceptive text were to generalize across different datasets. Finally, motion words generally denote concrete events, and therefore contradict the general finding of deception lacking in detail (H3).
It is therefore relatively unsurprising that, aside of reduced first-person pronoun use and emotional load, Newman et al.'s \cite{Newmanetal2003} results do not resurface in other studies.
%

Based on a review of prior research, Hancock et al. \cite{Hancocketal2008} formulated seven hypotheses on linguistic deception cues:

\begin{itemize}
\item[``(a)] Liars will produce more words during deceptive conversations than during truthful conversations.
\item[(b)] Liars will ask more questions during deceptive conversations as compared to truthful conversations.
\item[(c)] Liars will use fewer first-person singular but more other-directed pronouns in deceptive conversations than in truthful conversations.
\item[(d)] Liars will use more negative emotion words during deceptive conversations than during truthful conversations.
\item[(e)] Liars will use fewer exclusive words and negation terms during deceptive conversations as compared to truthful conversations.
\item[(f)] Liars will avoid causation phrases during deceptive interactions relative to truthful interactions.
\item[(g)] Liars will use more sense terms during deceptive interactions as compared to truthful interactions.'' \cite{Hancocketal2008}
\end{itemize}

The test subjects were divided between \emph{motivated} and \emph{unmotivated} liars based on whether the experimenter had provided false information (later revoked) about the importance of the ability to lie for success in life. Some hypotheses received confirmation from all liars (a--c, g), some only from motivated liars (f), and others from neither (d, e). Hypothesis (c) is indicative of a more general property of deception: the lack of a personal narrative (H4). However, a contrasting result is provided by the confirmation of (g): the increase of sense-related terminology. Sensation indicates a personal narrative, making this result contrast with the more general finding that deception tends to correlate with the lack of first-hand knowledge and detail (H3).

Lee et al. \cite{Leeetal2009} tested the ability of various linguistic features to predict deception in data from 30 deceptive and 30 truthful participants answering questions. While their initial hypothesis contained eight conglomerate properties, only one was statistically significant: \emph{certainty}, as calculated with a five-feature proxy measure comprised of causation words (e.g. \emph{because}, \emph{hence}), insight words (e.g. \emph{think}, \emph{know}), certainty words (e.g. \emph{always}, \emph{never}), first-person singular pronouns, present-tense verbs, and tenacity verbs (e.g. \emph{is}, \emph{has}). All five predicted deception in a statistically significant manner. These results are partly contradictory with Hancock et al's \cite{Hancocketal2008}, who found that (motivated) liars tended to avoid causation phrases. Further, the increase of first-person pronouns contrasts with many other studies, where their high use has correlated negatively with deception (H4).

\begin{table}
  \captionsetup{singlelinecheck=off}
  \begin{center}
    \begin{tabular}{|p{1cm}|p{4cm}|p{6cm}|p{1.1cm}|p{1.1cm}|}
    \hline
    \textbf{Study} & \textbf{Test setting} & \textbf{Deception cues} & \textbf{Support} & \textbf{Oppose} \\
    \hline
    \cite{Burgoonetal2003} \cite{Zhouetal2004a} &
    A theft-based game \cite{Burgoonetal2003}; a variant of the Desert Survival Problem \cite{Zhouetal2004a, Lafferty:Eady74} &
     
    quantity, \textbf{reduced immediacy}, expressiveness, informality, \textbf{affect}, reduced complexity, reduced diversity, \textbf{reduced specificity} & H1, H3 & \\
    
    \hline
    
    \cite{Zhouetal2004b} &
    Two variants of the Desert Survival Problem \cite{Lafferty:Eady74} &
    verbs, modifiers, word length, punctuation, modal verbs, individual reference, group reference, \textbf{emotiveness},
    content diversity, redundancy, \textit{perceptual information}, \textit{spatiotemporal information}, errors,
    \textbf{affect}, imagery, pleasantness, positive activation, positive imagery, negative activation & H1 & H3 \\
    \hline
    
    \cite{Newmanetal2003} &
    Reported views about abortion, friendship, and a mock crime scenario. &
    \textbf{reduced first person pronouns}, \textit{reduced third person pronouns}, reduced exclusive words, \textbf{negative
    emotion words}, \textit{motion words} & H1, H4 & H4 \\
    \hline
    
    \cite{Hancocketal2008} &
    Conversations between two participants &
    quantity, questions, \textbf{reduced first person singular pronouns}, \textbf{other-directed pronouns}, \textit{sense
    terms} & H4 & H4 \\
    \hline
    
    \cite{Leeetal2009} &
    A questioner-responder game & 
    causation words, \textbf{insight words}, \textbf{certainty words}, \textit{first-person singular pronouns}, present-tense verbs, tenacity verbs & H2 & H4 \\ \hline
    
    \cite{Mihalcea:Strappava2009} &
    Reported views about abortion, capital punishment, and friendship &
    \textbf{reduced self-related words}, \textbf{certainty-related words} & H2, H4 & \\ \hline
    \end{tabular}
    
    \caption[deception_experimental_caption]{Linguistic cues of experimentally elicited deception \\
     \textbf{Bold}: support H1--H4 \\
     \textit{Italics}: do not support H1--H4}
    \label{tab:deception_experimental}
  \end{center}
\end{table}

Mihalcea and Strapparava \cite{Mihalcea:Strappava2009} classified truthful and deceptive opinions concerning political and personal issues (abortion, capital punishment, and the responder's best friend) gathered via Amazon Mechanical Turk. At best they achieved a $70 \%$ accuracy with a Na\"ive Bayes classifier. They report a decrease of self-related words and an increase of certainty-related words as indicative of deception. Both results are in line with general findings of deception typically instantiating attempted persuasion (H1--H2) and a lack of first-hand experience (H3--H4).

In summary, the studies reviewed in this section suggest certain common features of experimentally elicited deception, but also include some unclear and even contradictory results. The results are collected in Table \ref{tab:deception_experimental}. The table additionally shows which of the hypotheses H1--H4 receive support or are contradicted by the findings.
A general trend is visible: deceivers often try to artificially emphasize what they say by using emotional and certainty-related terminology (H1--H2), while not providing detailed information about the topic they address (H3). However, contradictory results exist especially with respect to features related to the first-person narrative (H4). Many studies also support some of the hypotheses but oppose others. For instance, in Hancock et al.'s \cite{Hancocketal2008} data deception correlated both with reduced first-person pronoun usage and increased sense-terminology. The first of these features supports H4, but the latter points to the opposite direction, as sense-terminology often relates to descriptions of first-hand encounters. A similar case is found in Newman et al. \cite{Newmanetal2003}, who detected both reduced first-person pronoun usage and increased motion-word usage as indicators of deception.

\subsubsection{Non-elicited deception}
\label{sec:deception-non-elicited}

We now move on to deception in texts which have not explicitly been requested to be deceptive. These include both real-world corpora, as well as texts produced in experimental conditions where deceptiveness was not asked but was later evaluated based on independent criteria.

A common dataset for real-life deception has been the Enron e-mail corpus \cite{Klimt:Yang2004}.\footnote{\url{http://www.cs.cmu.edu/~enron/}} Keila and Skillicorn \cite{Keila:Skillicorn05} detected deceptive text from the Enron corpus, using features drawn from Zhou et al. \cite{Zhouetal2003} and Newman et al. \cite{Newmanetal2003} on the linguistic cues of deception: reduced use of first and third person pronouns and exclusive words, and increased use of negative emotion words and motion words. As discussed in Section \ref{sec:deception_elicited}, assuming these features to always indicate deception is not unproblematic. Further, while Keila and Skillcorn's manual evaluation indicated that the e-mails ranked high by these properties contained deceptive e-mails, the lack of ground truth makes it impossible to properly evaluate their results. Keila and Skillicorn additionally note that not only deception but other ``marked'' types of communication were also indicated by these features, such as  otherwise inappropriate messages.

Louwerse et al. \cite{Louwerseetal2010} predicted fraud in the Enron corpus using a five-point abstractness scale based on prior work by Semin and Fiedler \cite{Semin:Fiedler91}, who classified verbs and adjectives on the following scale, (a) being the most concrete and (e) the most abstract (examples from Semin and Fiedler \cite{Semin:Fiedler91}):

\begin{itemize}
\item[(a)] Descriptive Action Verbs: \textit{hit}, \textit{yell}, \textit{walk}
\item[(b)] Interpretative Action Verbs: \textit{help}, \textit{tease}, \textit{avoid}
\item[(c)] State Action Verbs: \textit{surprise}, \textit{amaze}, \textit{anger}
\item[(d)] State Verbs: \textit{trust}, \textit{understand}
\item[(e)] Adjectives: \textit{distraught}, \textit{optimal}
\end{itemize}


Louwerse et al. \cite{Louwerseetal2010} further divided adjectives to four analogical classes based on (a)--(d). Using Semin and Fiedler's \cite{Semin:Fiedler91} assessment that abstractness indicates low verifiability and low informativity, they predicted that high abstractness would correlate with deception. The email database was divided into sixteen events based on sending times, some of which were highly correlated with deception taking place within the Enron corporation. Regression analysis demonstrated that these events correlated with linguistic cues of high abstractness, providing support for the hypothesis.

Additionally, based on the results of Newman et al. \cite{Newmanetal2003} and Hancock et al. \cite{Hancocketal2008} (see Section \ref{sec:deception_elicited}), Louwerse et al. \cite{Louwerseetal2010} further investigated the correlation of deceptive events in the Enron corpus with first and third person pronouns, causal adverbs, negation, the connective ``but'', and email length. Of these, first person pronouns and negations were partially indicative of deceptive events, but the results were \textit{contrary} to the prior studies \cite{Newmanetal2003, Hancocketal2008}, as first person pronouns were used \emph{more} in deceptive emails rather than less.

Larcker and Zakolyukina \cite{LarckerZakolyukina12} studied linguistic properties of fraudulent and truthful financial statements by Chief Executive Officers (CEOs) and Chief Financial Officers (CFOs) in conference calls.
Their results diverged significantly between CEOs and CFOs. Differences were found in e.g. negations and extremely negative emotion words, which correlated positively with deception for CFOs but not CEOs. Some cues were even contrastive, as deception correlated with certainty-related words among CFOs, but hesitation-related words among CEOs. One possible reason for these differences could be that the features reflect the personal style of the CEOs/CFOs themselves rather than their deceptiveness. However, some commonalities were found: deceptive CEOs and CFOs both used more general group references, less non-extreme positive emotion terms and less third-person plural pronouns. While the prevalence of general group references indicates distance and thus supports H4, the other indicators seem particular to this study, as they are not replicated in other studies. They also bear no clear relation to H1--H4.

Toma and Hancock \cite{Toma:Hancock12} compared the linguistic properties of fraudulent and truthful online dating profiles. While the profiles were written in experimental settings, deception was not encouraged, and was only detected by comparing the profiles to ground-truth gathered about the users. Deception correlated significantly with reduced first-person singular pronouns, increased negations, a lower word count, and a decrease in negative emotion vocabulary. While the last feature stands in opposition to many other studies \cite{Zhouetal2003, Newmanetal2003, Keila:Skillicorn05}, it is unsurprising considering the context: a deceptive dating-profile would most likely exaggerate positive qualities and downplay negative ones. Hence, it is unlikely that this result would generalize across different text types.

Crabb \cite{Crabb2014} used POS-tags and lexical diversity for deception detection from the Enron corpus. She used two methods: clustering with the Expectation-Maximum algorithm, and calculating means for each feature in isolation to detect statistically significant differences with respect to deception-cues identified in prior research \cite{Zhouetal2003, Zhouetal2003b, Hancocketal2008, Afrozetal2012, Leeetal2009, Fengetal2012, Keila:Skillicorn05, Louwerseetal2010}.
Two clusters were deemed most relevant due to the high occurrences of modal, base and present tense verbs, second-person pronouns, and function words. However, while emails in these clusters generally had higher values for such features than those in other clusters, not all such values were statistically significant. Further, the lack of ground truth in the Enron corpus prevented any conclusive inferences to be made concerning the prevalence of deception in the clusters.
%

Hancock and Markowitz \cite{Hancock:Markowitz2014} used linguistic information to classify papers by the social psychologist Diederik Stapel, who famously fabricated data to many publications. They observed the following tendencies in Stapel's fraudulent papers in comparison to truthful ones:

\begin{itemize}
\item more terms related to scientific methodology
\item more amplifying terms (e.g. \textit{extreme}, \textit{exceptionally}, \textit{vastly})
\item more certainty-related terminology
\item more emotional terminology
\item fewer diminisher terms (e.g. \textit{somewhat}, \textit{partly}, \textit{slightly})
\item fewer adjectives
\end{itemize}

Hancock and Markowitz' \cite{Hancock:Markowitz2014} results thus provide support for the hypotheses that deceivers generally exaggerate the content they want the receiver to believe (H1) and their level of certainty (H2), while providing less qualitative descriptions (H3). Their model correctly classified $71 \%$ of Stapel's papers. While this was significantly better than random choice, the authors express caution about the feasibility of their method for broader forensic use, citing the large error rate and the domain-specificity of scientific discourse.

\begin{table}
  \begin{center}
    \begin{tabular}{|p{1cm}|p{4cm}|p{6cm}|p{1.1cm}|p{1.1cm}|}
    \hline
    \textbf{Studies} & \textbf{Data} & \textbf{Deception cues} & \textbf{Support} & \textbf{Oppose} \\  
    \hline
    
    \cite{Louwerseetal2010} &
    Enron e-mails \cite{Klimt:Yang2004} &
    \textbf{abstractness}, negations, \textit{first person pronouns} & H3 & H4 \\ \hline
    
    \cite{LarckerZakolyukina12} &
    Conference call transcripts &
    \textbf{general group references}, reduced non-extreme positive emotion terms, reduced third-person plural pronouns & H3 &
    \\ \hline
    
    \cite{Toma:Hancock12} &
    Online dating profiles &
    \textbf{reduced first-person singular pronouns}, negations, reduced word count, \textit{reduced negative emotion words} &
    H4 & H1 \\ \hline

    \cite{Hancock:Markowitz2014} &
    Fraudulent scientific papers &
    words related to scientific methodology, amplifying terms, \textbf{certainty-related
    words}, \textbf{emotional words}, reduced diminisher terms, reduced adjectives & H1, H2 & \\ \hline
    
    \cite{Crabb2014} &
    Enron e-mails \cite{Klimt:Yang2004} &
    modal, base and present tense verbs, \textbf{second-person pronouns}, function words & H4 & \\ \hline
    
    \end{tabular}
    \caption{Linguistic cues of non-elicited deception \\
    \textbf{Bold}: support H1--H4 \\
    \textit{Italics}: do not support H1--H4}
    \label{tab:deception2}
  \end{center}
\end{table}

Studies on non-elicited deception are summarized in Table \ref{tab:deception2}. The results are mostly in line with experimental research (Table \ref{tab:deception_experimental}): common features include high emotional load (H1), certainty-related terminology (H2), abstractness (H3), and the reduced use of first-person pronouns (H4). However, as in experimental studies, the evidence is contradictory concerning emotional words and first-person pronouns.
\subsection{Deception detection from online text}
\label{sec:deception-security}

In this section we focus on two specific topics relevant for online security: \textit{fake reviews} and \textit{troll comments}. We argue that both present unique properties not inferrable from the results reviewed in Section \ref{sec:deception-general}. We further discuss alternative methods for their detection, and evaluate the importance of pure text analysis as a tool for these tasks.

\subsubsection{Fake reviews}
\label{sec:fake-reviews}

One major source of deceptive online text is \textit{fake reviewing}, where the reviewer deliberately attempts to (mis)lead the audience into believing something about a product \cite{Wangetal2012, Rinta-Kahila:Soliman2017}.
Fake reviews may have special properties in comparison to other forms of deception, and are therefore allocated a separate section in this survey. As Yoo and Gretzel \cite{Yoo:Gretzel2009} point out, fake reviewers are often professionals, and can typically model their writing on real reviews.
%
Additionally, a fake review does not need to be fraudulent with respect to the author's actual opinions. Instead, its deceptiveness stems from the \textit{purpose} of the author to spam a site for some ulterior reason instead of providing informative reviews. Hence, fake reviews are not necessarily disbelieved by the author, but the content is irrelevant to the author's true goal: they instantiate \textit{deception of intention}.

For supervised methods, obtaining labeled data constitutes a major challenge, and studies have typically collected their own data. The largest corpus of elicited fake reviews has been compiled by Ott et al. and contains 400 fake and 400 truthful reviews of both the positive \cite{Ottetal2011} and negative \cite{Ottetal2013} kind.\footnote{The corpus is available at \url{http://myleott.com}.} Additionally, the website Yelp provides a corpus of filtered reviews suspected to be fake.\footnote{\url{https://www.yelp.com/dataset}}

\noindent{\textbf{Human written reviews}}
Ott et al. \cite{Ottetal2011} detected deceptive pieces in TripAdvisor hotel review data generated via Amazon Mechanical Turk. Combining psycholinguistic features from the LIWC software \cite{Pennebakeretal2007} and word bigrams, they achieved an accuracy of $89.8\%$, and summarize their results as follows:

\begin{quote}
``(...) truthful opinions tend to include more sensorial and concrete language than deceptive opinions;
(...) we observe an increased focus in deceptive opinions on aspects external to the hotel being reviewed \\
(...) our deceptive reviews have more positive and fewer negative emotion terms. \\
(...) we find increased first person singular to be among the largest indicators of deception'' \cite{Ottetal2011}
\end{quote}

These findings stand in stark contrast to H4, since here deception is indicated by an \textit{increase} in first-person pronouns and hence a more personal narrative. This trend turns out to be prevalent in fake reviews, providing support for Yoo and Gretzel's \cite{Yoo:Gretzel2009} contention that fake reviews differ from other forms of deception. On the other hand, Ott et al. also found that fake reviews were more abstract and less specific, in line with H3 and against Yoo and Gretzel's analysis of fake reviews having a special status due to the availability of information.

Feng et al. \cite{Fengetal2012} further improved Ott et al.'s \cite{Ottetal2011} results by adding syntactic phrase structure to the stylometric evaluation of the same dataset, reaching $91.2\%$ accuracy. As features they used both word bigrams and abstract syntactic relations derived from a context-free grammar parse.

Ott et al.'s first study \cite{Ottetal2011} was concerned with \textit{positive} hotel reviews. In a subsequent study \cite{Ottetal2013}, the same authors applied the method to \textit{negative} reviews, also gathered via Amazon Mechanical Turk. They achieved a F-score of c.a. $86 \%$ with n-gram-based support vector machines (SVMs). Negative fake reviews contained less spatial information and had a larger verb-to-noun ratio than truthful reviews. They also manifested an excess of negative emotion terms, in direct contrast with the high use of positive terms in the prior study. Ott et al. interpret these results as opposing the hypothesis that negative words indicate the emotional distress involved in lying \cite{Ekman:Friesen68}. Rather, the increased use of emotional terminology can be explained by the intention of the deceiver to communicate certain contents, which is why the prevalent emotions will vary along with these intentions.
High emotional load may still be a useful deception cue, but it results from a more general property of \textit{emphasis}, and is not ubiquituously negative.

Yoo and Gretzel \cite{Yoo:Gretzel2009} tested seven hypotheses on the linguistic properties of fake hotel reviews:

\begin{itemize}
\item[``(a)] Deceptive reviews contain more words.
\item[(b)] Deceptive reviews are less complex.
\item[(c)] Deceptive reviews are less diverse.
\item[(d)] Deceptive reviews contain less self-references (immediacy).
\item[(e)] Deceptive reviews contain a greater number of references to the hotel brand.
\item[(f)] Deceptive reviews contain a greater percentage of positive words.
\item[(g)] Deceptive reviews contain a smaller percentage of negative words.''
\end{itemize}

Hypotheses (e--g) were confirmed, while (a--d) were not. In fact, the opposite hypotheses to (b) and (d) received support: fake reviewers used more complex language and more self-references than truthful reviewers. These results imply that fake reviews may differ from other types of deception by often being conducted by experts. Further, based on Ott et al.'s results \cite{Ottetal2011, Ottetal2013}, it seems likely that (f--g)'s success was due to the reviews' promotional nature, and would plausibly not be replicated on negative review data.

Hue et al. \cite{Huetal2012} base their analysis of deceptive reviews on two properties: \emph{sentiment} and \emph{readability}. Sentiment is relevant since fraudulent reviewers likely have the intention of slanting the review either in favour of or against the product. Hue et al. further argue that in addition to sentiment varying randomly across different reviews by a genuine author, the same should be true of \emph{readability}, measurable by e.g. the Automated Readability Index (ARI) based on the amount of characters within words and the amount of words within sentences \cite{Smith:Senter67}. In contrast, they maintain that readability should remain high and consistent across fraudulent reviews, since these aim at a maximally general audience. Using the Wald-Wolfowitz Runs test to detect non-randomness in manually labelled data from Amazon reviews, they provide empirical confirmation for constancy in both sentiment and readability as indicators of fake reviews.

Li et al. \cite{Lietal2014} studied linguistic generalities across fake reviews, which they divided between expert-generated and crowdsourced spam. They note that the commmon assumption of fake reviews lacking in detail \cite{Ottetal2011, Lietal2013} is not true of expert-generated reviews. For crowdsourced reviews, their results accorded with previous studies indicating that fake reviews are less specific, and thus contain less descriptive terms like nouns or adjectives \cite{Buller:Burgoon1996a, Bulleretal1996, Biberetal1999, Raysonetal2001}. This, however, was not the case for expert-generated fake reviews, which were highly informative and descriptive. Other linguistic cues Li et al. discovered were exaggerated sentiment and the overuse of first person singular pronouns. The latter result was contradictory to many previous studies proclaiming that deceivers avoid talking about themselves \cite{Knapp:Comaden1979, Bulleretal1996, Newmanetal2003, Zhouetal2004b}.

Xu et al. \cite{Xuetal2015} based their unsupervised fake review classifier on the text's \textit{generality}, i.e. lack of informativity. The model ranked reviews based on ``spamicity'', the top reviews being most spam-like. Based on Ott et al.'s claim that online review sites typically contain $8\% - 15\%$ spam, they tested their model by treating the top $k$ \% as spam, where the value of $k$ was varied between $5 \%$, $10\%$ and $15\%$. Accuracy was tested by comparing the top $k$ \% to its supervised classification by SVMs \cite{Chang:Lin2011}. Applying the model to three review datasets, Xu et al. obtained F-scores of $75.2 \% - 78.8 \%$ with $k=5\%$, $72.2 \% - 76.6 \%$ with $k = 10 \%$, and $69.4 \% - 71.7 \%$ with $k = 15 \%$. As they note, their method only works for reviews for products that are unavailable for the fake reviewer to investigate, such as restaurants or hotels. The assumption of fake reviews lacking specificity does not hold for products of which much information is available via commercials or other descriptions, since the reviewer could use such information in constructing the spam \cite{Yoo:Gretzel2009}.

\begin{table}
  \begin{center}     
    \begin{tabular}{|p{1cm}|p{4cm}|p{6cm}|p{1.1cm}|p{1.1cm}|}
    \hline
    \textbf{Studies} & \textbf{Data} & \textbf{Deception cues} & \textbf{Support} & \textbf{Oppose} \\ 
    \hline
    
    \cite{Yoo:Gretzel2009} &
    Hotel reviews (positive, experimentally elicited) &
    high complexity, \textit{first person pronouns}, brand names, positive words, decreased negative words & & H4 \\
    \hline
    

    \cite{Ottetal2011} &
    Hotel reviews (positive, crowdsourced) &
    \textbf{reduced specificity}, \textbf{external information}, positive sentiment, reduced negative sentiment, \textit{first person singular pronouns}, high verb-to-noun ratio & H3 & H4 \\
    \hline
    
    \cite{Ottetal2013} &
    Hotel reviews (negative, crowdsourced) &
    (in negative reviews:) \textbf{reduced specificity}, negative emotion terms, high verb-to-noun ratio & H3 & \\
    \hline
    
    \cite{Huetal2012} &
    Amazon.com reviews &
    high readability, constancy of sentiment & &
    \\ \hline
    
    \cite{Lietal2014} &
    Hotel, restaurant, and doctor reviews (crowdsourced) &
    unspecificity (non-expert reviews), \textit{specificity} (expert reviews), \textbf{exaggerated sentiment}, \textit{first person singular pronoun} & H1 & H3, H4 \\    
     \hline
     
     \cite{Xuetal2015} &
     Amazon audioCD, TripAdvisor (hotels), Yelp (restaurants) &
     \textbf{text generality} & H3 & \\
     \hline
    \end{tabular}
    \caption{Linguistic properties of fake reviews \\
    \textbf{Bold}: support H1--H4 \\
    \textit{Italics}: do not support H1--H4}
    \label{tab:fake-reviews}
  \end{center}
\end{table}


The results from fake review studies are summarized in Table \ref{tab:fake-reviews}.
A recurring theme is the lack of specificity, but this depends on the assumption that the reviewer does not access information about the product \cite{Xuetal2015}. High or low sentiment has also been demonstrated to be relevant as in other forms of deception, but its direction depends on the nature of the review (positive or negative). Increased use of the first person pronoun stands in contrast to results received on other forms of deception (see Section \ref{sec:deception-general}), supporting Yoo and Gretzel's \cite{Yoo:Gretzel2009} contention about fake reviews constituting a \textit{sui generis} type of deception. Yoo and Gretzel's analysis of fake reviews being special due to the amount of detailed information available receives partial support from Li et al. \cite{Lietal2014}, but only for expert-generated reviews.

\noindent{\textbf{Automatically generated reviews}}
\textit{Automatic text generation} is a vast field within NLP \cite{Dale:Reiter2000}, and poses an additional threat to review sites.
Detecting automatically generated reviews is a different task than detecting man-made fake reviews, due to the different nature of the deception. In automatically generated reviews, the deception concerns \textit{identity}: the message is meant to look like it is written by a human, but is in fact machine-generated.

Hovy \cite{Hovy2016} automatically generated fake reviews using a 7-gram Markov chain trained with data from the review site Trustpilot. For classification, he used logistic regression with word n-grams ($1 \leq n \leq 4$) as features. The classifier additionally sought irregularities between age, gender, review category, and n-grams. Adding such meta-information to the model significantly improved its ability to fool the classifier. However, while exact copies of training reviews were removed, a 7-gram model will likely reproduce large chunks of the training data. Duplicate or similarity detection between the training data and the generated reviews was not conducted by Hovy.

Yao et al. \cite{Yaoetal2017} generated fake reviews with a character-level Recurrent Neural Network (RNN) trained with restaurant reviews from Yelp.\footnote{\url{https://www.yelp.com/dataset}}
They were unable to distinguish RNN-generated reviews reliably from those in the Yelp corpus, using linear SVMs with various linguistic features, the plagiarism detection method Winnowing \cite{Schleimeretal2003}, or human evaluators from Amazon Mechanical Turk ($n = 594$). These results demonstrate that machine-generated fake reviews can resist classification by common methods. However, Yao et al. suggest an alternative defence against their RNN-generated fake reviews, based on statistical differences in character distributions between generated reviews and the training corpus.

Juuti et al. \cite{Juutietal2018} utilized Neural Machine Translation (NMT) to generate context-appropriate restaurant reviews. They demonstrated a superior performance to Yao et al. \cite{Yaoetal2017} in fooling human users trying to distinguish between genuine and generated reviews. In their user study, Juuti et al. were able to avoid detection at a rate of $3.5/4$, as opposed to $0.8/4$ with Yao et al.'s method. By controlling the context (e.g. restaurant name, type of food, review rating etc.) they can further generate reviews of a specific type with a single NMT model. Despite successfully deceiving human readers, they were able to detect generated reviews with a very high F1-score of $97 \%$, using an AdaBoost classifier trained on words, POS n-grams, dependency tag n-grams, and NLTK's \cite{Bird:Loper2004} readability features.

Recent developments in automatic text generation demonstrate that automating the task of fake reviewing is an increasing threat. As generated reviews do not display particular similarities to human-written fake reviews \cite{Juutietal2018}, there is no reason to believe that the hypotheses H1--H4 have any particular relevance here. Text generators mimic the writing style of their training corpus, which by assumption contains mostly genuine reviews. Hence, standard deception detection has no bearing on this issue. Instead, classifying generated reviews requires knowledge of the generation model itself, in which case they remain detectable \cite{Yaoetal2017, Juutietal2018}. However, as such knowledge is not always available, the problem cannot be considered solved.


%
\subsubsection{Trolling and cyberbullying}
\label{sec:trolls}

\textit{Troll users} deliberately post malicious content to online forums, either to harass others for amusement or with the intention of advancing an agenda. \textit{Paid trolls} post professionally on behalf of an institution (e.g. a political candidate, government, or corporation), while \textit{mentioned trolls} are identified as trolls by other users \cite{Mihaylov:Nakov2016}.
\textit{Cyberbullying}\hypertarget{link:cyberbullying}{ }is a related phenomenon, where the author targets a particular victim instead of an entire forum. While trolls or cyberbullies are not exclusively dishonest, there is major overlap in the purposes of a deceiver and a troll: both write content with a purpose other than its truthful communication. Especially professional trolls have no necessary connection between their actual opinion and what they write, and therefore are likely to write content they believe to be false.
Additionally, even if a troll writes something he believes, his \textit{intention} is nevertheless fraudulent.
Similar considerations apply for cyberbullying.
It is therefore initially plausible that trolling/cyberbullying and other forms deception detection might overlap in linguistic features.

Cambria et al. utilized \textit{sentic computing} to classify texts according to the likelihood of being authored by a troll \cite{Cambriaetal2010}. As a knowledge-based method, sentic computing is more grammatically and semantically oriented than many other current NLP approaches, as it is built on a pre-programmed set of ``common-sense'' concepts and inference patterns \cite{Cambria:Hussain2015}. Cambria et al. used the method to attest the emotional content of the data, based on the following scales:

\begin{itemize}
\item[``1.] the user is happy with the service provided (Pleasantness)
\item[2.] the user is interested in the information supplied (Attention)
\item[3.] the user is comfortable with the interface (Sensitivity)
\item[4.] the user is disposed to use the application (Aptitude)'' \cite{Cambriaetal2010}
\end{itemize}

Cambria et al., classify human emotions by these four dimensions together with \textit{polarity}, i.e. whether the emotion is positive or negative. They found that troll posts had a high absolute value of Sensitivity and a generally negative polarity. Trolls manifested either significantly high or low levels of comfort with the interface, together with a negative sentiment. Testing on a manually classified test set of troll and non-troll Twitter messages, Cambria et al. received an F-score of $78\%$ ($82\%$ precision, $75\%$ recall).



J.-M. Xu et al. \cite{Xuetal2012b} used sentiment analysis to detect cyberbullying from Twitter. They manually classified seven emotions relevant for bullying: anger, embarrassment, empathy, fear, pride, relief and sadness. Fear was by far the most common emotion in their cyberbullying dataset, whether the author was identified as a bully, a victim, an accuser or a bystander. These results indicate that fear-related terminology may be informative of bullying as a \textit{topic}, but not of the status of the author as the bully.

Troll posts are commonly negative, being targeted against some viewpoint or a person. Using the hypothesis that negative sentiment is indicative of trolling, Seah et al. \cite{Seahetal2015} applied sentiment analysis to online forum posts. They received a generalized receiver operating characteristic of $78 \%$ with binary classification and $69 \%$ with ordinal classification.

%
Mihaylov and Nakov \cite{Mihaylov:Nakov2016} used various linguistic and metalinguistic features to detect both paid and mentioned troll comments in news community forums. They received an F1-score of $78 \%$ for mentioned trolls and $80 \%$ for paid trolls. Despite the slight differences between the troll types, they conclude that both paid and mentioned trolls behave similarly in comparison to non-trolls. Among linguistic features, bag-of-words fared well overall, as opposed to more abstract grammatical properties like POS-tags. However, metalinguistic features were more effective than any linguistic feature.


\begin{table}
  \begin{center}
    \begin{tabular}{|p{1cm}|p{4cm}|p{6cm}|}
    \hline
    \textbf{Studies} & \textbf{Data source} & \textbf{Cues to trolling/bullying} \\
    \hline
    
    \cite{Cambriaetal2010} &
    Twitter &
    negative sentiment \\
    \hline
    
    \cite{Seahetal2015} &
    Discussion forum &
    negative sentiment \\
    \hline
    
    \cite{Mihaylov:Nakov2016} &
    Discussion forum &
    bag-of-words, negative sentiment \\
    \hline
    

    \cite{Chenetal2012} &
    Youtube comments &
    offensive words, intensifiers \\
    \hline
    

    \cite{Xuetal2012b} &
    Twitter &
    fear-related words \\
    \hline
    

    \end{tabular}
    \caption{Linguistic properties of troll comments}
    \label{tab:trolls}
  \end{center}
\end{table}

\textit{Offensive language} is an important factor in trolling and cyberbullying. Following Jay and Janschewitz \cite{Jay:Janschewitz2008}, Chen et al. \cite{Chenetal2012} characterize offensive language as vulgar, pornographic or hateful. For evaluating the overall offensiveness of sentences, Chen et al. used an offensive word lexicon that included manual measures for words collected from Youtube comments, and further measurements based on a word's syntactic context. Detecting offensive users in an online discussion corpus, they receive $78 \%$ in both precision and recall.

A related issue is \textit{hate speech}\hypertarget{link:hate-speech}, the detection of which has been explored in a number of studies \cite{Dinakaretal2012, Gitarietal2015, Burnap:Williams2015, Nobataetal2016, Waseemetal2016, Schmidt:Wiegand2017, Badjatiya2017, Davidsonetal2017, Wulczynetal2017, Zhangetal2017}.
Hate speech is only occasionally deceptive, which is why we do not discuss it in detail here. However, a brief summary of the findings in this field is worth taking into account. First, character n-grams have generally performed well across hate speech datasets, which is likely due to their flexibility across spelling variants \cite{Mehdad:Tetreault, Schmidt:Wiegand2017, Grondahletal2018}. Second, offensive word lexicons have not performed well in the absence of n-gram features \cite{Nobataetal2016, Schmidt:Wiegand2017}. Third, while deep learning approaches have become more popular than more basic machine learning methods \cite{Zhangetal2017, Badjatiya2017}, a comparative study by Gr\"{o}ndahl et al. \cite{Grondahletal2018} demonstrated that their performance did not significantly differ when trained on the same datasets. Finally, the same study showed that even state-of-the-art approaches are highly vulnerable to simple text transformations like removing spaces or adding innocuous words. Such evasion techniques are similar to earlier methods of evading spam detection \cite{Lowd:Meek2005, Zhouetal2007}.

Reviewing the main results of the troll detection studies discussed in this section, the most prevalent cue is \emph{negative sentiment}. It clearly does not suffice, as non-troll messages can also have negative sentiment, and not all trolls are negative. At most the results indicate that negative sentiment is indicative of an increased probability of trolling. Like with fake reviews, studies on troll and cyberbullying detection reflect the fact that \textit{content}, much more than writing style, has determined the success of classification. Hence, the results do not support the plausibility of a content-invariant detection scheme.
%
%
\subsection{Deception detection: future prospects}
\label{sec:deception-future-prospects}

Summarizing the studies reviewed in Sections \ref{sec:deception-general}--\ref{sec:deception-security}, some results have been replicated in multiple studies. In particular, deceptive texts often have a high emotional load and a large frequency of certainty-related terms, while troll posts tend to have a negative sentiment. However, a more fine-grained analysis demonstrates that the relevant features have been highly content- and context-sensitive. Hence, they are unlikely to scale beyond the semantic domains of particular datasets.
%
Therefore\hypertarget{link:deception-suggestions}, we suggest that detecting deception is more efficient with methods outside of purely linguistic analysis. Specifically, we recommend \textit{semantic comparison} between different documents.
For example, Mihaylov et al. \cite{Mihaylovetal2015} used various measures to detect troll users from an online news community forum, among which was \textit{comment-to-publication similarity}. Their hypothesis was that trolls may be prone to deliberately cite news articles in a misleading fashion to support their own perspective. This feature had a positive effect on classification, and links troll-detection to \textit{rumor-debunking}, where similar content-comparison methods prevail \cite{Rubin2017}.


A related approach to unsupervised fake review classification is the detection of semantic and grammatical similarities between reviews.
Such methods rely on the assumption that spammers tend to repeat the same message in multiple places.
Narisawa et al. \cite{Narisawaetral2007} classified spam based on the similarity measure of \emph{string alienness}, obtaining F1-scores between $50 \%$ and $80 \%$.
Uemura et al. \cite{Uemuraetal2011} detected review spam using \textit{document complexity} (based on the amount of similar documents within the corpus), and received F1-scores between $66\%$ and $73\%$.
%
Lau et al.'s \cite{Lauetal2011} unsupervised model was also based on duplicate detection based on semantic overlap.

Of course, semantic comparison measures do not detect author intentions, such as deceptiveness. This task may well be 	\textit{impossible in principle} if only text data is used. Our literature review indicates that deceptiveness as an author intention does not leave a content-invariant linguistic trace. Deception may, at most, correlate with certain linguistic properties in particular semantic domains. Restricted to a domain, linguistic features may still be useful in aiding deception detection, at least when used in combination with metalinguistic data concerning e.g. user behavior on the forum.
\section{Author identification and adversarial stylometry}
\label{sec:author-identification}

In this section we discuss \textit{author identification} from an \textit{adversarial} perspective, where detection and its evasion are treated as competing tasks.
Avoiding deanonymization or profiling involves obfuscating writing style, for which a variety of techniques has been suggested.
Style transformation for anonymization or imitation purposes constitutes a type of \textit{deception}, albeit different in kind from those reviewed in Section \ref{sec:deception}. There, we characterized deception, broadly understood, as attempting to lead the reader into believing something false. In style transformation, the relevant information involves \textit{author identity} or \textit{profile}, the first concerning individual identity and latter membership in a broader group.
Unless mentioned otherwise, the studies reviewed have concerned author identification. With respect to profiling, features are likely to vary depending on the classes under interest (age, gender, occupation etc.), which makes results less generalizable. However, some style transformation studies have concentrated on profiling instead of identification \cite{Prabhumoyeetal2018, Shettyetal2018}.

We begin by reviewing the state-of-the-art in stylometry research in Section \ref{sec:author-identification-general}. We then advance to information security -related uses of author identification, to which we devote Section \ref{sec:deanonymization}. After introducing the \textit{deanonymization attack} \cite{Narayanetal2012}, we dedicate Section \ref{sec:adversarial_stylometry} to discussing its mitigation by style obfuscation or imitation.

\subsection{Author identification}
\label{sec:author-identification-general}

The success of author identification depends on the validity of the \textit{Human Stylome Hypothesis} (HSH) \cite{vanHalterenetal2005}, which maintains that authors have a unique writing style that is retained to a significant extent between different texts, even across variation in semantic content. Its validity is obviously not a binary matter, and will inevitably differ between authors and datasets. Nevertheless, general trends found in empirical work contribute useful indicators of its suitability for real-world applications.
In this section we provide a concise review of existing work in author identification. For further discussion, we refer to prior surveys dedicated solely to this topic \cite{Stamatatos2009, Nealetal2017}.

There is a close affinity between writing styles and \textit{idiolects} as speaker-specific (mental) grammars, which can differ among members of the same language community.
The idea that lexical repositories and grammatical rules vary between individual speakers is a well-attested linguistic fact \cite{Sapir27, Bloch48, Coulthard2004, Otheguyetal2015}. While this gives the HSH initial plausibility, it is worth bearing in mind that idiolects reflect a large variety of factors, not limited to choices between content-equivalent stylistic variants. Indeed, the linguistic literature on idiolects has often focused on \textit{semantic} variation between authors \cite{Fokkema:Ibsch87, Louwerse2004}. Another problem for the HSH is the prevalence of \textit{style-shift}. As the sociolinguist William Labov stated: ``There are no single style speakers'' \cite{Labov84}. If a speaker can change between styles in different contexts, stylometric classification might not capture author identity but rather ``style clusters'' spanning many authors, who conversely can belong to multiple clusters. Recent results on large-scale stylometric clustering accord with this hypothesis \cite{Nealetal2017}.

The problematicity of HSH notwithstanding, concrete examples of author identification can be found outside academic research. In 2011, an American man was found to be the true author of a blog supposedly written by a Syrian woman \cite{Bennett2011}. While stylometry was not responsible for the finding, Afroz et al. \cite{Afrozetal2012} demonstrated that a close linguistic correlation could be found between the blog and other texts by the same author. In 2013, stylometric analysis performed by Peter Millican and Patrick Juola on the novel \textit{The Cuckoo's Nest} revealed its likely author to be J.K. Rowling under a pseudonym, which she later confirmed.\footnote{For Juola's description of the study, see http://languagelog.ldc.upenn.edu/nll/?p=5315.} Juola \cite{Juola2013} also reports a real-life court case where an asylum-seeker claimed to have written newspaper articles critical of his government, for which he would have faced persecution if not granted the asylum. As evidence, he provided other articles provably written by him, and the court had to evaluate their similarity to the contested articles. In such cases, stylometry can provide assistance for making decisions with large-scale consequences.

A significant problem in the field is the lack of consensus on which features to use \cite{Rudman1998, Rudman2010, Ramyaaetal2004, Juola2012}. \hypertarget{link:writeprints}{The} most prevalent collection argued to be optimal for identifying individual authors even from short texts is the ``Writeprints'' feature set \cite{Zhengetal2006, Abbasi:Chen2008}. It consists of a variety of character-based, lexical, syntactic, and structural features, as partly presented in Table \ref{tab:Writeprints}. The set was introduced by Zheng et al. \cite{Zhengetal2006}, who used it to identify authors with $97.69 \%$ accuracy from a corpus containing 20 candidate authors, and $30-92$ articles of $84-346$ words from each candidate.
I has since been used in multiple studies \cite{Abbasi:Chen2008, Afrozetal2012, McDonaldetal2012, Farkhundetal2013, Almisharietal2014, Overdorf:Greenstadt2016}, and is partially implemented in the JStylo software \cite{McDonaldetal2012}.

\begin{table}
  \begin{center}
    \begin{tabular}{|lc|p{4cm}|}
    \hline
    
    \textbf{Feature types} & & \textbf{Example features} \\
    \hline

    \multirow{2}{*}{Lexical} &
    \multicolumn{1}{|c|}{Character} &
    number of characters,
    number of letters,
    number of digits,
    frequency of letters,
    frequency of special letters
    \\ \cline{2-3} & \multicolumn{1}{|c|}{Word} &
    number of words,
    average word length,
    vocabulary richness,
    average sentence length \\
    
    \hline
    
    Syntactic & &
    frequency of punctuations,
    frequency of function words \\
    
    \hline
    
    Structural & &
    number of sentences,
    number of paragraphs,
    number of sentences/words/characters in a paragraph,
    has quotes \\
    
    \hline
    
    Content-specific & &
    frequency of content specific keywords \\
    
    \hline
    
    \end{tabular}    

    \caption{Examples of the Writeprints features (270 altogether) \cite{Zhengetal2006}}
    \label{tab:Writeprints}
  \end{center}
\end{table}

While large-scale comparisons between different features applied to the same datasets have been rare, existing comparative studies indicate that \textit{low-level features} like short character n-grams (including unigrams) have a systematically high performance.
Grieve \cite{Grieve2007} applied $39$ features prevalent in prior work (before 2007) to a single dataset, using the chi-square test for producing a ranking of the most likely authors. The top-$5$ feature types with the best performance were word unigrams (including punctuation) and character n-grams in the order $2 > 3 > 4 > 1$ from most to least successful. In contrast, positional features, vocabulary richness, sentence length, and word length had only modest or poor performance. A higher prevalence of function words in comparison to content words further improved the success rate, which is in line with traditional assumptions of style being especially manifested in function words \cite{Mosteller:Wallace1964a}.

Juola \cite{Juola2012} summarizes over $3$ million experiments he and colleagues made on the same datasets comparing combinations of features, pre-processing methods, and classifiers included in the authorship attribution software JGAAP \cite{Juola2009}. The datasets were taken from an author attribution competition \cite{Juola2004}, and are provided with JGAAP. The best results were achieved with punctuation features, using nearest neighbours with Manhattan distance for analysis. According to Juola, a likely explanation of these results is that the corpus exhibited a particularly large variance in quotation marks and other non-alphanumeric notation. Therefore, the results are not applicable to datasets where such features have been normalized.

Potthast et al. \cite{Potthastetal2016b} evaluate the performance of $15$ suggested techniques on three datasets. Their results suggest that using \textit{compression} improves the stability of performance across different corpora. The basic idea behind compression is that single compressed files are produced of the candidate author's texts both alone and together with the unknown author's texts, and divergence is then measured between these files \cite{Kukushkina2001, Martonetal2005}. Potthast et al. further remark that character features were the most effective overall. Similar conclusions regarding compression and character features are reached in a larger comparative study by Neal et al. \cite{Nealetal2017}, who evaluate $14$ open-source algorithms on a corpus containing $1000$ authors. Summarizing their results, the authors note that low-level features like characters fare better especially on smaller samples, where high-level features like syntactic dependencies are sparse.

In addition to the discrepancy between feature sets across different studies, a further problem in stylometry research concerns whether the features are more indicative of \textit{style} or \textit{content}. Evidently, highly content-related features like lexical choices are not applicable across different genres or domains \cite{Argamonetal2009, Nealetal2017}. This is likely among the main reasons for the success of function words \cite{Mosteller:Wallace1964a, Burrows87, Zhao:Zobel2005}, which have also been argued to correlate with personality types \cite{Chung:Pennebaker2007}, and form the basis of the linguistic profiling software LIWC \cite{Pennebakeretal2007, Tausczik:Pennebaker2010}. Small susceptibility to content changes is also a virtue of punctuation features \cite{Juola2012} and grammatical structure \cite{Hirst:Feiguina2007, Raghavanetal2010, Posadas-Duranetal2014}.

With respect to classification algorithms, most stylometry research has focused on traditional supervised machine learning methods, such as SVMs, decision trees, Bayesian classification, or distance metrics \cite{Stamatatos2009, Nealetal2017}. SVMs have been particularly popular due to their strong performance on high-dimensional and sparse data \cite{Stamatatos2009}. Deep learning applications have recently become more prominent, with a particular focus on recurrent and convolutional neural networks \cite{Bagnall2015, Ge:Sun2016, Surendranetal2017}. Brocordo et al. \cite{Brocardoetal2017} also experiment with \textit{deep belief networks}, which belong to the class of probabilistic generative models. While deep learning methods have generally demonstrated a strong performance in many NLP tasks \cite{Goldberg2016, Youngetal2017}, their large training data requirements present problems with smaller author corpora \cite{Nealetal2017}. Recent approaches to \textit{transfer learning} in NLP have attempted to improve classifier scalability by first training an initial model to perform some task using a large training set, and subsequently fine-tuning the model for different tasks with smaller additional training sets \cite{Howard:Ruder2018, BERT2018}. However, the transferability of other text classifiers to author identification is yet to be studied.

\hypertarget{link:large-scale-identification}{In} order to succeed beyond artificial experimental conditions, author identification should be feasible across a large number of candidates with small example corpora from each. However, as Neal et al. \cite{Nealetal2017} state in their survey, existing techniques face challenges in such settings. To detect potential author groups, they performed graph-based clustering in a large corpus. The number of clusters ($16$) was much smaller than the number of authors ($1000$), and there was no clear separation between authors. Neal et al. note the possibility that the clusters represent ``meta-classes'' characterizing multiple authors, and a single author can belong to many such classes. These results are in line with Labov's dictum that a speaker is never bound to a single style, and vice versa \cite{Labov84}. However, it is worth noting that if these hypothetical ``meta-classes'' are simply assimilated with the clusters, the claim is difficult to either confirm or falsify.

The largest attempt at author identification so far has been contributed by Narayanan et al. \cite{Narayanetal2012}, whose data was derived from $100,000$ different blogs. The features used were post length, vocabulary richness, word shape (the distribution of lower- and upper-case letters), word length, and the frequencies of letters, digits, punctuation, special characters, function words, and syntactic category pairs. Narayanan et al. correctly predicted the author in over $20 \%$ of the cases, which is a significant increase from random chance. Still, from the perspective of deanonymization, the approach cannot be considered successful, as it was far more likely to yield a false prediction than the correct one.

There is no single universally accepted protocol for author identification that could be used directly ``out-of-the-box''. While significant overlap can be found in the features and classifiers used, most studies have been unique with respect to the particular subset of features, and have not conducted systematic evaluations between different combinations. Software like Signature,\footnote{\url{http://www.philocomp.net/humanities/signature.htm}} JGAAP \cite{Juola2009}, JStylo \cite{McDonaldetal2012}, and RStylo \cite{Ederetal2016} have been developed to alleviate this problem by allowing researchers to conduct stylometric tests with a simple GUI, selecting from a list of pre-programmed features and classifiers. Some of these systems are very restricted in the range of features they offer, which limits their application potential. For instance, RStylo only uses word- or character n-grams, and does not allow their combination in the same test. The most featurally sophisticated application is JStylo, which contains a subset of the Writeprints feature set \cite{Zhengetal2006}.

\hypertarget{link:character-ngrams}{Neal et al.} \cite{Nealetal2017} give two plausible reasons for the commonly observed effectiveness of short character n-grams in comparison to high-level properties like abstract grammatical relations. First, the latter are sparse in short texts, whereas all texts contain characters. Second, character-features are less susceptible to noise, such as misspellings or grammatical errors. In addition to these benefits, we believe that character-features can have an exceptionally high correlation with many other features. For instance, the prevalence of particular function words will impact the frequency of their characters, making character-features indirectly responsive to changes in function word use. Hence, low-level features like character n-grams have the potential to record (partial) information of a large variety of textual properties. They can therefore be expected to fare generally better than high-level features, at least with small corpora. Character-features also have the advantage of being \textit{language-independent} in the sense of requiring no language-specific pre-processing, such as tokenization, POS-tagging, or parsing \cite{Nealetal2017}.

\subsection{Implications for security and privacy}
\label{sec:deanonymization}

Author identification and profiling have a multifaceted relation to information security. Forensic studies have been on the forefront in traditional stylometry research, providing assistance in uncovering the identities of criminals \cite{McMenamin:Choi2002, Coulthard2004}. Similar methods can help to unmask \textit{troll users} in online forums. As a case study, Galn-Garcia et al. \cite{Galn-Garciaetal2014} linked troll profiles to their true profiles, and successfully applied the method to a real-life cyberbullying case. Another important application is the detection of \textit{doppelg\"{a}ngers} or \textit{sockpuppets}, i.e. users with multiple accounts.
Solorio et al. \cite{Solorioetal2013} used SVMs with $239$ linguistic features to detect sockpuppet accounts from Wikipedia user comments, and reached a $68 \%$ accuracy.
Afroz et al. \cite{Afrozetal2014} used stylometric techniques to link doppelg\"{a}nger users with unsupervised clustering, achieving $85 \%$ precision and $82 \%$ recall on an underground forum dataset.

In contrast to the assistance that author identification can provide for increasing online security, it also constitutes a \textit{privacy threat} by making it possible to deanonymize authors against their will.
Brennan et al. \cite{Brennanetal2011} propose an adversarial scenario they call \textit{Alice the anonymous blogger versus Bob the abusive employer}, where an employer uses stylometry to uncover the author of an anonymous complaint.
%
%
Another potential adversarial purpose of deanonymization is \emph{bullying} or \emph{harassment} \cite{Almisharietal2014}.
In general, the abusive part can be played by any person or institution, such as a government, corporation, or individual.
%
Narayanan et al. \cite{Narayanetal2012} coined the term \textit{deanonymization attack} to denote such scenarios. They further take their empirical results on blog author identification to indicate that the attack is not only a theoretical possibility but a real-life concern.

As we reviewed in Section \ref{sec:author-identification-general}, Narayanan et al. were able to detect a blog author from $100,000$ candidates with ca. $20 \%$ accuracy \cite{Narayanetal2012}. We also noted that, while these results demonstrate a major increase from random chance, they nevertheless most often fail to find the correct author. Still, they are genuinely disconcerting from the perspective of potential victims of a deanonymization attack.
In designing secure systems, it is essential to assume a low bar for the attacker and a high bar for the defender. Applying this principle to the deanonymization attack, we conclude that since stylometry can significantly increase the chances of the attacker correctly guessing the author's identity, it constitutes a genuine privacy threat.
Additionally, results on smaller author corpora ($\leq 20$) indicate that high accuracy can be achieved with features contained in Writeprints \cite{Zhengetal2006, Abbasi:Chen2008}, or similar sets \cite{Stamatatos2009, Nealetal2017}. In many variants of the deanonymization attack, assuming a fairly restricted candidate set is justified: for example, in Brennan et al.'s \cite{Brennanetal2011} scenario (see above) the candidates are restricted to Bob's employees.
%
%

Motivated by their findings, Narayanan et al. \cite{Narayanetal2012} recommend the development of automated tools for transforming writing style while preserving meaning.
The field of \emph{adversarial stylometry} \cite{Brennanetal2011} involves the study of such countermeasures to deanonymization.
In Section \ref{sec:adversarial_stylometry} we review the work conducted in this field, and evaluate whether the deanonymization attack can realistically be mitigated using existing methods.

\subsection{Adversarial stylometry}
\label{sec:adversarial_stylometry}


We define \textit{style obfuscation} as any method aimed at fooling stylometric classification. A more restricted variant of obfuscation is \textit{imitation}, where misclassification is intended to target a particular author. Imitation also constitutes an attack, with the original author as the attacker and the imitated author as the victim.
We divide obfuscation methods into three basic types: manual, computer-assisted and automatic. The subsections \ref{sec:manual-obfuscation}--\ref{sec:automatic-obfuscation} are divided by these methods, and within each subsection studies are reviewed in the order of publication. If one study has used several methods, its results are divided among the subsections.

\subsubsection{Manual obfuscation}
\label{sec:manual-obfuscation}

Brennan and Greenstadt \cite{Brennan:Greenstadt2009} experimented with two manual methods of masking the original author of a text: \textit{obfuscation} and \textit{imitation}. The former involved conscious altering of a text to avoid displaying properties characteristic of the author, and in the latter authors attempted to mimic the style of another writer. The results form the \textit{Brennan-Greenstadt Corpus}. In a subsequent study, Brennan et al. \cite{Brennanetal2011} used Amazon Mechanical Turk to crowdsource the obfuscation task. The results, along with the original corpus, form the \textit{Extended Brennan-Greenstadt corpus}, which is provided with the JStylo software \cite{McDonaldetal2012}.
Brennan et al. evaluated obfuscated and imitated texts with three methods: neural networks with the Basic-9 feature set,\footnote{\label{fn:Basic-9}The Basic-9 feature set consists of the number of unique words,
lexical density, Gunning-Fog readability index, character count (without whitespace),
the average number of syllables per word, sentence count, the average sentence length, and the Flesch-Kincaid Readability Test. Like Writeprints \cite{Zhengetal2006}, Basic-9 is also implemented in JStylo \cite{McDonaldetal2012}.} a synonym-based classifier \cite{Clark:Hannon2007}, and SVMs with the Writeprints features \cite{Zhengetal2006}. Both obfuscation and imitation resulted in the success rates of all methods dropping significantly, only the SVM-Writeprints classifier remaining above a chance level. Imitation also succeeded in reaching the correct targets. The SVM method was most resistant against both obfuscation and imitation. The effectiveness of the (original) Brennan-Greenstadt corpus against the authorship attribution program JGAAP \cite{Juola2009} was further demonstrated by Juola and Vescovi \cite{Juola:Vescovi2010}.
Amazon Mechanical Turk was also successfully used by Almishari et al. \cite{Almisharietal2014} to reduce automatic author recognition. Both obfuscation and readability evaluation were crowdsourced. On a scale from $1$ (``Poor'') to $5$ (``Excellent''), the average readability score was $4.29$, indicating success in retaining the original meaning to a significant degree.
%

The results reviewed here indicate that writing style can be manually altered to deceive author identification. Contrary to the strong interpretation of the HSH, it thus seems possible to change one's writing style, at least with deliberation.
However, manual obfuscation is very time-consuming and laborious. Having to consciously alter the style of everything one wants to write anonymously is not a scalable solution. Crowdsourcing is a possible way to outsource manual obfuscation, but Almishari et al. \cite{Almisharietal2014} note, sending your original writings to strangers constitutes a privacy risk. Conceivably, the adversary could even act as a Mechanical Turk worker and see the original text as a job offered by the author. Crowdsourcing is also relatively slow and costly to use.

\subsubsection{Computer-assisted obfuscation}
\label{sec:computer-assisted-obfuscation}

The idea of computer-assisted manual style obfuscation was introduced by Kacmarik and Gamon \cite{Kacmarik:Gamon2006}, who automatically evaluated the feature changes needed to make classification fail with Koppel and Schler's \cite{Koppel:Schler2004} author identification technique. They present a graph linking the features requiring modification, allowing the user to monitor their success at obfuscation.
Anonymouth \cite{McDonaldetal2012} uses the stylometric framework JStylo to evaluate a text written by the user against reference corpora. Based on this evaluation, it gives the user instructions on modifying the text to evade JStylo.
Day et al. \cite{Dayetal2016} developed the concept of Adversarial Authorship, and implemented it as an application called AuthorWeb. It displays a user other texts similar to their current writing in style, allowing them to obfuscate text by controlling which texts their current writing resembles.

Computer-assisted methods can be useful in comparison to the fully manual approach, given that they reduce the cognitive load involved in deciding which features to alter. Without external cues the author would need to guess which changes to make, which would be unattainable in practice. However, a problem with automatic evaluators like Anonymouth is their reliance on specific corpora and classifiers, which may be unavailable for the author. Furthermore, while computer-assisted obfuscation is easier than purely manual methods, choosing the candidate remains in the hands of the user.
Manual work is still needed, which makes these approaches unfeasible in a large scale.

\subsubsection{Automatic obfuscation}
\label{sec:automatic-obfuscation}

Fully automated author masking is, unsurprisingly, significantly more difficult to implement than manual or computer-assisted obfuscation. This section reviews studies conducted in this field, and evaluates the performance of the most prominent methods.
%

\noindent{\textbf{Iterative language translation}} The easiest and most common method used for automatic style alteration and hence author obfuscation is machine translation (MT) across one or more languages back to the source language. This can also be called Iterative Language Translation (ILT) \cite{Macketal2015}, as its effect can be increased by multiple iterations. It was first suggested by Rao and Rohatgi \cite{Rao:Rohatgi2000} and has since been applied in multiple studies, with divergent results. Here, we survey and analyze this research. All the studies reviewed have used English as the language to obfuscate.
With the exception of Keswani et al. \cite{Keswanietal2016}, Google Translate has been used for translation, sometimes compared with the Bing translator. It should also be noted that, with one exception \cite{Shettyetal2018}, all the studies were conducted prior to Google's major update that replaced phrase-based statistical MT with deep neural networks (DNNs) \cite{GoogleTranslate2016}.

ILT was attempted by Brennan et al. \cite{Brennanetal2011} with both Google and Bing translators. They used German and Japanese as intermediate languages, experimenting with English--German--English, English--Japanese--English, and English--German--Japanese--English (the first two exemplifying \textit{one-step} and the last \textit{two-step} translation). Success of obfuscation was measured with a synonym-based classifier, a neural network with the Writeprints features, and an SVM with the Basic-9 features (see footnote \ref{fn:Basic-9}). Approximately, in one-step translations the success rate of the synonym-based method declined from $85 \%$ to $75 \%$, and the success rates of the neural network and SVM classifiers declined from $65 \%$ to $50 \%$. Google obfuscated more with Japanese and Bing with German as the intermediate language. With the exception of Bing translations evaluated with neural networks and SVMs, two-step translation did not markedly improve obfuscation. Overall, the results were deemed underwhelming, and the authors concluded that state-of-the-art MT in 2009 did not provide sufficient means for author obfuscation.

Caliskan and Greenstadt \cite{Caliskan:Greenstadt2012} also used Google and Bing's translators with German and Japanese as intermediate languages, but with English--Japanese--German--English as the two-step translation order. The success of obfuscation was measured with JGAAP \cite{Juolaetal2006}\cite{Juola2009} and JStylo \cite{McDonaldetal2012}, using what they call the Translation Feature Set, which was selected via optimization from the Basic-9 and Writeprints feature sets \cite{Brennan:Greenstadt2009, Zhengetal2006}.\footnote{The Translation Feature Set contained the following features: average characters per word, character count, function words, letters, punctuation, special characters, top letter bigrams, top letter trigrams, words, and word lengths \cite{Caliskan:Greenstadt2012}.}
After obfuscation, the average recognition rate remained high at $92 \%$, which accorded with Brennan et al.'s \cite{Brennanetal2011} pessimistic conclusions about ILT.
Caliskan and Greenstadt further classified translated texts based on the translator (Google or Bing) with an average success rate of $91 \%$, indicating that the translation algorithm itself can be ``fingerprinted'' if appropriate stylometric features are used.

Using Google Translate, Almishari et al. \cite{Almisharietal2014} reduced the linkability between the translated text and the original author by increasing the amount of intermediate languages up to nine, randomly drawn from the $64$ languages offered by Google Translate in 2014. They conducted a readability review of $60$ translations (produced via nine intermediate languages) via Amazon Mechanical Turk, receiving an average score of $2.8 / 5$. The readability of a subset of translated texts was further improved manually (also via Mechanical Turk), retaining author anonymity. However, without a comparison to texts produced with other methods, the readability score alone does not tell much about the status of the obfuscated texts.

Mack et al. translated English blog texts back and forth through Arabic, Chinese and Spanish with one to three iterations  \cite{Macketal2015}. The results were evaluated with four Author Identification Systems (AISs): a unigram-based AIS, the O. de Vel et al. AIS \cite{DeVeletal2001}, a combination of the previous two called Hybrid-I, and Hybrid-I with added  syntactic features (Hybrid-II). A genetic algorithm called Genetic and Evolutionary Feature Selection (GEFeS) was further used to mask nonsalient features from each AIS to improve their performance. The addition of GEFeS resulted in Hybrid-II having the best performance overall on recognizing the author from the non-obfuscated test corpus ($52 \%$), the other AISs having rates of c.a. $20 \%-25 \%$. ILT lowered the identification rate of all AISs with all intermediate languages, Arabic always faring the best. The most significant result was the decrease of Hybrid-II plus GEFeS' identification rate from 51.65 \% to c.a. 10 \% with all languages on the first iteration, and below 10 \% with Arabic. Further iterations did not markedly change the identification rates, irrespective of the language or the AIS.

Day et al. \cite{Dayetal2016} used Google Translate with Spanish and Chinese as intermediate languages. Mack et al.'s Hybrid-II AIS \cite{Macketal2015} was used to evaluate the success of obfuscation. With the first iteration, accuracy dropped from $54 \%$ to $6 \%$ with Spanish and to $10 \%$ with Chinese. Further iterations did not decrease the identification accuracy, as the second iteration resulted in $6 \%$ with Spanish and $11 \%$ with Chinese, and the third iteration yielded $7 \%$ with Spanish and $11 \%$ with Chinese. Day et al. further used Latent Semantic Analysis (LSA) \cite{Dumais2004} to analyse semantic overlap between the original and obfuscated texts. This algorithm gives a value between -1 (no similarities) and 1 (full similarity). The LSA-values were $0.86$ for the Spanish-mediated text and $0.77$ for the Chinese-mediated text. Day et al. further fingerprinted the intermediate language with the JGAAP software \cite{Juola2009}, receiving accuracies of $93 \%$ (Spanish) and $90 \%$ (Chinese) on the first iteration, $98 \%$ (Spanish) and $97 \%$ (Chinese) on the second iteration, and $98 \%$ (Spanish) and $99 \%$ (Chinese) on the third iteration. The number of iterations was also fingerprintable, although less accurately than the translator.

Keswani et al. \cite{Keswanietal2016} applied ILT to the author masking task arranged by the PAN 2016 digital forensics event. Using Moses \cite{Koehnetal2007}, they created their own translation model trained with the Europarl corpus \cite{Koehn2005}. The text was translated through German and French. Three features were evaluated of the obfuscated texts \cite{Potthastetal2016}. \textit{Safety} indicates how well the obfuscated text manages to hide original authorship, and was measured by the obfuscation's impact on classification by various author verification systems from previous PAN tasks. Keswani et al.'s method succeeded in obfuscation $25 \% - 42 \%$ of the time, depending on the dataset. The \textit{sensibility} of the obfuscated text and its \textit{soundness}, i.e. similarity in meaning with the original text, were both manually evaluated from a small subset of texts. Keswani et al.'s text was, in Potthast et al.'s words, considered ``impossible to read or understand'' by the PAN 2016 evaluator due to the frequency of errors \cite{Potthastetal2016}.

As a baseline for evaluating their Generative Adversarial Network (GAN) approach called A$^4$NT (discussed below), Shetty et al. \cite{Shettyetal2018} applied four variants of ILT with Google Translate, using German, French, Spanish, Finnish, and Armenian as intermediate languages between two and five iterations. None of the variants significantly reduced the classification rate on a word-based Long Short Term Memory (LSTM) network, the largest drop being from $90 \%$ to $81 \%$ in F1-score. Shetty et al.'s user study also indicated that ILT did not succeed in maintaining semantic similarity.


\begin{table}
  \begin{center}
  \captionsetup{justification=centering}
    \begin{tabular}{|c|c|c|c|c|}
    \hline
    \textbf{Study} & \textbf{Translator(s)} & \textbf{Languages} & \textbf{Iterations} & \textbf{Success} \\
    \hline

    \cite{Brennanetal2011} & Google, Bing & German, Japanese & $1-2$ & No \\

    \hline
    
    \cite{Caliskan:Greenstadt2012} & Google, Bing & Japanese, German & $1-2$ & No \\

    \hline
    
    \cite{Almisharietal2014} & Google & (Random) & $\leq$ $9$ & Yes \\
    
    \hline
    
    \cite{Macketal2015} & (Not told) & Arabic, Chinese, Spanish & $1-3$ & Yes \\  
    
    \hline
    
    \cite{Dayetal2016} & Google & Spanish, Chinese & $1-3$ & Yes \\
    
    \hline
    
    \cite{Keswanietal2016} & Moses & German, French & $2$ & Unclear / No \\
    
    \hline
    
    \cite{Shettyetal2018} & Google (DNN) & German, French, Spanish, Finnish, Armenian & $2-5$ & No \\ \hline
    
    \end{tabular}
    \caption{A comparison of studies using ILT for style obfuscation \hfill \\ (``Success'' = the reported success of the approach in deceiving author identification, based on the source paper.)}
    \label{tab:MT-obfuscation}
  \end{center}
\end{table}

Based on our review, one reason for the differing outcomes in ILT-obfuscation seems to be the languages used. As summarized in Table \ref{tab:MT-obfuscation}, studies have generally used different intermediate languages and numbers of iterations.
While small-scale comparisons have been made, the effects of varying the languages have not been systematically evaluated. Results on the effects of iterations are also indecisive. Almishari et al. \cite{Almisharietal2014} decreased identification accuracy by adding iterations, whereas Mack et al. \cite{Macketal2015} and Day et al. \cite{Dayetal2016} did not. More systematic comparative research would be needed to properly evaluate the effects of the languages, the number and direction of iterations, and the translation method. With respect to the last, it is possible that Shetty et al.'s \cite{Shettyetal2018} failure to obfuscate with Google Translate even across five intermediate languages was affected by its update from a statistical algorithm to a DNN \cite{GoogleTranslate2016}.

It is also likely that ILT will decrease the grammaticality and hence readability of the text, and/or differentiate its semantic interpretation from the original text \cite{Rao:Rohatgi2000, Nealetal2017}. Successful change of style would require three properties from the resulting text: (i) grammatically soundness, (ii) retention of the original meaning, and (iii) evasion of author identification. These properties have not been properly measured \textit{together} in the ILT-obfuscation studies reviewed in this section.
Additionally, ILT is unreliable due to not taking into account the \textit{direction} of the obfuscation. The changes might take the classification to \textit{any} direction, which may or may not aid obfuscation. As Shetty et al. \cite{Shettyetal2018} note, a style obfuscation system should ideally only enact changes that take classification to the desired direction, and no more.
%
%
%

\noindent{\textbf{Rule-based substitutions}}
Khosmood and Levinson \cite{Khosmood:Levinson2008} outline a basic model of rule-based style imitation based on grammatical changes. The purpose of the system is to alter the style of a source text until it is maximally close to that of a target corpus. First, both the source text and the target corpus are analysed based on \textit{style markers}, which are predetermined linguistic features. Next, the styles of the source text and the target corpus are compared, and their stylistic distance is determined based on some metric calculated from the  style markers. If the distance is large enough, a predetermined modification rule is applied to the source text that alters some of its style markers. A comparison between the altered source text and the target corpus is made, resulting in a finished transformation if their stylistic distance is close enough, and to another modification otherwise. This \textit{Classification-Transformation Loop} (CTL) \cite{Khosmood:Levinson2009} is continued until the stylistic distance is sufficiently close or no more transformations are available.

Khosmood and Levinson \cite{Khosmood:Levinson2008} applied the CTL to a US Department of Justice memorandum excerpt, with a part of Orwell's Animal Farm as the target corpus. They used 10 style markers for analysis and comparison, and modified the source text with three transformation rules of de-hyphenation, lexical substitution and acronym expansion. Stylistic distance was measured with the root-mean-square-error value, which reduced from $5.77$ in the source text to $5.63$ in the modified text.
In another study \cite{Khosmood:Levinson2010}, the same authors divided text into n-grams between one and five units, and searched for synonyms for each from Wordnet \cite{Miller1995, Fellbaum1998}. For each word/phrase, one synonym was chosen above others based on its commonness in the target corpus or another database, and the original word/phrase was replaced with it. Using JGAAP for evaluation, Khosmood and Levinson succeeded in obfuscating seven out of thirteen texts. Readability was manually evaluated by the authors as ``correct'', ``passable'' or ``incorrect'', concluding that the majority of texts were ``passable''. As no additional evaluators were used, and the authors only provide two example obfuscations, these results are difficult to assess.

Mansoorizadeh et al. \cite{Mansoorizadehetal2016} employed Wordnet-based lexical substitution for the PAN 2016 Author Masking Task \cite{Potthastetal2016}. Synonyms for the top $200$ words used by the author of the training text were found from Wordnet via NLTK. Two criteria were used in choosing the replacement synonym from the alternatives provided by Wordnet: Wu et al's semantic similarity metric \cite{Wu:Palmer1994}, and the occurrence probability in the original word's context measured with a 4-gram language model trained with the Brown corpus. In their evaluation of the PAN 2016 Author Masking Task, Potthast et al. \cite{Potthastetal2016} note that Mansoorizadeh et al.'s algorithm is very conservative, changing at most one word per sentence. While retaining readability well, this also results in reduced safety against author identification. Further, certain replacements resulted in semantic errors, such as \textit{machine} being exchanged for \textit{car}. The system succeeded in obfuscation in $14 \% - 25 \%$ of cases depending on the dataset. In Potthast et al.'s manual sensibility evaluation, the obfuscated text received a grade of $2/5$ on a scale from $1$ (excellent) to $5$ (fail), mainly due to punctuation errors. In a further manual evaluation of similarity to the original text on a three-point scale of ``incorrect'', ``passable'' and ``correct'', all obfuscated texts were graded as ``correct'' or ``passable''.

In the same PAN 2016 Author Masking Task, Mihaylova et al. applied various alterations by replacing elements with others, merging, splitting, removal etc. \cite{Mihaylovaetal2016}. In Potthast et al.'s evaluation \cite{Potthastetal2016}, Mihaylova et al. received the best results of all PAN 2016 Author Masking Task contenders in safety (i.e. the success of obfuscation), with an average impact of 36 \% -- 49 \% depending on the dataset. However, in the manual evaluation of sensibility and soundness, the obfuscated texts were deemed practically unreadable and semantically odd. In a subsequent study, the same authors applied similar but improved techniques to the same test setting, shifting stylistic features toward their average distribution in the training set \cite{Karadzhovetal2017}. Using multiple author identification methods from prior PAN competitions, their method achieved an average accuracy drop between $10 \%$ and $16 \%$, and maintained a superior readability compared to their prior method.

In terms of retaining the original meaning, rule-based substitution is a more secure obfuscation method than ILT, as it allows deterministic user control of the output. Especially with grammatical changes, transformations can be limited to have only minor semantic impact. However, the scalability of hand-crafted rules across a large variety of datasets is difficult to attain \cite{Shettyetal2018}.
With lexical replacements semantic retention is harder to control, as the appropriateness of paraphrases can be highly  context-dependent. If WordNet is used for synonym replacement, context effects can partly be accounted for by using sense disambiguation techniques, such as the Lesk algorithm \cite{Lesk86, Ramakrishnanetal2004}. WordNet represents words in the uninflected lemma format, which restricts synonym replacement to contexts where the surface form is identical with the lemma. The Paraphrase Database (PPDB) \cite{Ganitkevitchetal2013} is the major alternative to WordNet, and links inflected forms directly. Derived from parallel corpora used for MT, PPDB also involves information about the appropriate syntactic environment for the paraphrases. However, since the phrases are represented as raw text, it does not directly allow the use of Lesk or other semantic sense disambiguation algorithms.

\noindent{\textbf{MT between styles}}
In addition to translating across different languages, MT can be used within the same language to automatically paraphrase text. Importantly, such a method could allow not only obfuscation of the original author but automatic \textit{imitation} of a predetermined style. MT was used for style transformation by Xu et al. \cite{Xuetal2012}, who paraphrased Shakespeare as modern English and evaluated the results both manually and with three automatic methods based on cosine similarity (n-gram overlap), language models with Bayesian probability estimation, and logistic regression. The automatic metrics correlated with human judgement to a significant degree.

In addition to using ILT for obfuscation (see above), Day et al. \cite{Dayetal2016} applied iterative paraphrasing, creating the paraphrase dataset with the online tool Plagiarisma. Paraphrasing decreased the author identification rate with Hybrid-II \cite{Macketal2015} from $54 \%$ to $7 \%$ in the first iteration, $1 \%$ with the second iteration, and $6 \%$ with the third iteration. The LSA-value for paraphrased text was $0.80$, indicating relatively high lexical overlap with the original text. Like the MT algorithms, paraphrasing itself was detectable, with fingerprinting accuracies of $86 \%$ on the first iteration, $91 \%$ on the second iteration, and $95 \%$ on the third iteration.

More recently, \textit{neural machine translation} (NMT) techniques \cite{Luongetal2015, GoogleTranslate2016} have been adopted for automatic style imitation. The input is first mapped to a style-neutral representation, and then a new sentence is generated from this representation while controlling target style. However, the transformations implemented in these studies have often involved semantic changes, as in altering sentiment or political slant \cite{Huetal2017, Shenetal2017, Prabhumoyeetal2018}. In contrast, the main goal of adversarial stylometry is to retain semantic content to a maximal extent while fooling the author classifier. This is evidently not achieved with examples like Prabhumoye et al.'s political slant transformation from ``i thank you, sen. visclosky'' to ``i'm praying for you sir'' \cite{Prabhumoyeetal2018}. Such examples may deceive a Democrat-Republican classifier, but they also change the original meaning too much to constitute viable forms of transformation for anonymization purposes. Since we are concerned with adversarial stylometry and not content alteration, we do not review research on the latter. An important aspect of future work is a more systematic application of the suggested methods to different kinds of tasks, with a particular focus on their ability to retain content across stylistic changes.

In addition to experimenting on political slant and sentiment, Prabhumoye et al. \cite{Prabhumoyeetal2018} also tackle the issue of gender profiling, which falls under our scope by being a purely author-related, non-semantic feature. The basis for their method is the notion that translation to another language will remove many style-specific features \cite{Rabinovichetal2016}. First, they train translators between English and French to both directions, and begin the style transformation process by translating the original sentence to French. They then process the French translation with the encoder part of the French-to-English translator. The decoder part of the translator is a generative model that takes the French encoding as a context vector and produces an English target sentence. They split this decoder into different variants, which are trained to produce sentences allocated to particular categories by a CNN classifier. The resulting sentences are thus the combined effect of the original French-English translator and the class-based tuning of the English decoder.
Prabhumoye et al. compare their back-translation method with Shen et al.'s \cite{Shenetal2017} cross-aligned autoencoder approach, which is similar but uses a different algorithm for generating the intermediate style-neutral representation. 
The gender classifier's original accuracy of $82 \%$ was reduced to $40 \%$ with cross-aligned autoencoders and $43 \%$ with back-translation.
In a manual fluency evaluation on $60$ random sentences, gender imitation by cross-aligned autoencoders received an average rating of $2.42/4$, while imitation by back-translation received $2.81/4$.

Shetty et al. \cite{Shettyetal2018} present a Generative Adversarial Network (GAN) -based approach to style transformation, which they title Adversarial Author Attribute Anonymity Neural Translation (A$^4$NT). A GAN consists of a classifier trained to discriminate between two or more classes, and a generative model that is trained to fool this classifier \cite{Goodfellowetal2014}. A$^4$NT is an unsupervised approach where an encoder-decoder network is trained to generate sentences which fool a word-based LSTM author classifier, but also maintain a maximal semantic proximity to the original sentence. Semantic retention was measured as a combination of two components: the probability of reconstructing the original sentence via a reverse A$^4$NT-transformation, and the distance of sentence embeddings constructed using a pre-trained embedding model \cite{Conneauetal2017}. A$^4$NT was tested across three classification tasks: blog author gender, blog author age, and political speeches by Barack Obama vs. Donald Trump. In all tasks, the method lowered classification accuracy to random chance or below. However, these results only concerned the same classifier as used in training the GAN. Shetty et al. further show that blog age classification F1-score is dropped from $87 \%$ to $62 \%$ with the best of $10$ alternative classifier candidates. Corresponding results from the two other tasks are not shown.
For assessing semantic similarity, they use the MT evaluation metric Meteor \cite{Denkowski:Lavie2014}, which measures n-gram overlap using additional paraphrase tables. They receive scores of $0.69$, $0.79$, and $0.29$ in the gender, age, and Obama/Trump tasks, respectively. Shetty et al. note that these results exceed those received with automatic paraphrasing methods \cite{Lietal2018}, although such comparison is problematic as the studies involve different corpora. Finally, a user-study indicated that human evaluators preferred A$^4$NT to ILT via Google Translate with a similar obfuscation success.

Of the approaches reviewed here, only A$^4$NT has a built-in mechanism for semantic retainment. In spite of this, even its example transformations often include drastic semantic changes, as seen in the following Obama-Trump transformations taken from Shetty et al. \cite{Shettyetal2018}:

\begin{itemize}
\item[] ``their situation is getting worse.'' $\rightarrow$ ``their media is getting worse.''
\item[] ``(...) because i do care'' $\rightarrow$ ``(...) because they don't care.''
\item[] ``that's how our democracy works.'' $\rightarrow$ ``that's how our horrible horrible trade deals.''
\end{itemize}

A system that cannot secure sufficient semantic retention is unreliable for real-life application, irrespective of its success in fooling author identification. Overall, recent advances in NMT and GANs show promise in generating stylistic transformations, but further research is required to evaluate the feasibility of such methods in more realistic scenarios against a large variety of classifiers.
Beyond adversarial stylometry, the transformation of writing style has been studied within \textit{automatic text simplification}  \cite{Siddharthan2010, Siddharthan2011, Wubbenetal2012, Narayan:Gardent2014}, which in turn belongs to the broader field of \textit{paraphrase generation} \cite{Madnani:Dorr2010, Lietal2018}. Effects of these methods on author obfuscation have yet to be investigated, but increasing the interaction between these fields would likely be beneficial to both sides.

\section{Conclusions}
\label{sec:Conclusions}

Section \ref{sec:intro} presented the following three questions concerning deception detection based on writing style:

\begin{itemize}
\item[Q1] Does deception leave a content-independent stylistic trace?
\item[Q2] Is the deanonymization attack a realistic privacy concern?
\item[Q3] Can the deanonymization attack be mitigated with automatic style obfuscation?
\end{itemize}

Based on the literature review conducted in Section \ref{sec:deception}, Q1 was answered negatively. We demonstrated that linguistic features that have correlated with deception have been too specific to particular semantic domains to constitute genuine stylistic ``deception markers''. The practical consequence of this finding is that stylometric analysis has plausible utility for deception detection only if the training and test domain are sufficiently similar. Furthermore, even when successful, stylistic markers of deception are likely to be \textit{content-based} correlates rather than indicators of general psychological mechanisms behind lying.

Our review suggest that alternative approaches to pure stylometry are likely more effective in detecting textual deception. These include, in particular, \textit{content comparison}, \textit{similarity detection}, and using \textit{metadata}. Content comparison allows detecting texts that contain claims with a pre-established truth-value based on an external knowledge base \cite{Rubin2017}. False claims are not deceptive if they are sincerely believed (see Section \ref{sec:deception}), but a strong correlation between falsity and deception is nevertheless likely. Surface-level similarity between texts has also proven helpful in finding trolls or spammers, who tend to repeat the same across many discussions  \cite{Narisawaetral2007, Uemuraetal2011, Lauetal2011}. Finally, information beyond linguistic content has been more effective in detecting fake reviews \cite{Mukherjeeetal2013, Rayana:Akoglu2015} or trolls \cite{Mihaylov:Nakov2016} than linguistic content. Stylometric classification can assist such techniques, but is severely limited as a stand-alone solution.

With respect to Q2, we argued that stylometry-based deanonymization constitutes a realistic privacy threat, especially if the set of potential authors is small (ca. $\leq 20$) \cite{Zhengetal2006, Brennan:Greenstadt2009, Brennanetal2011, Almisharietal2014}. Even though author identification has not proven sufficiently scalable to larger sets of authors (ca. $> 1000$) \cite{Nealetal2017}, stylometry can still significantly increase the likelihood of finding the correct author, even among $100 000$ candidates \cite{Narayanetal2012}. From the perspective of an author wishing to retain anonymity, these results are legitimately worrying. With the constant increase in the availability of corpora and computing power, the deanonymization attack will likely continue to be a growing privacy threat. We therefore consider the further development of automatic style obfuscation tools as not merely an academic excercise, but to have important real-life consequences for information security.

Turning to Q3, manual obfuscation remains potentially effective against the deanonymization attack \cite{Brennan:Greenstadt2009, Brennanetal2011, Almisharietal2014}, and tools like Anonymouth \cite{McDonaldetal2012} can help in this task. Fully automatic approaches, in contrast, suffer from the  difficulty of balancing sufficient obfuscation success with semantic faithfulness to the original text. So far, only simple rule-based approaches have allowed securing semantic retention, as transformations can be limited to semantically vacuous choices \cite{Khosmood:Levinson2008, Karadzhovetal2017}. However, these methods are very limited in application, and have not demonstrated sufficient obfuscation success.
Only one of all the studies reviewed in Section \ref{sec:automatic-obfuscation} included a semantic similarity measure in the algorithm \cite{Shettyetal2018}, and even it had trouble with too severe semantic alterations. Approaches have largely relied on \textit{a priori} assumptions about ILT or paraphrase replacement not altering semantics, which has not been sufficiently confirmed.
Additionally, while user studies are important for assessing readability and semantic retention, the lack of established baselines makes the results difficult to evaluate. Merely comparative measures between different techniques are also inadequate, as they do not demonstrate whether the transformations are acceptable, but only which are preferred under an obligatory choice.

The detectability of obfuscation methods themselves has not been sufficiently investigated, as only two of the studies we reviewed had conducted such an evaluation \cite{Caliskan:Greenstadt2012, Dayetal2016}. All ILT variants could be detected with a high accuracy in both studies, including even the number of intermediate languages. Day et al. also successfully fingerprinted a paraphrase-based MT-algorithm \cite{Dayetal2016}.
These results indicate that even if obfuscation succeeds, obfuscated texts could still be distinguished from original texts. However, it bears emphasis that such classification requires knowledge of the obfuscation algorithm, which may not be available. The general property of being obfuscated with \textit{any} method is unlikely to leave a stylistic trace.
The situation is similar to the case of automatically generated fake reviews (Section \ref{sec:fake-reviews}), where the detection of generated text is possible, but only provided that the generation algorithm is known \cite{Yaoetal2017, Juutietal2018}.

Summarizing our discussion on adversarial stylometry, while promising frameworks for automatic text transformation exist especially within NMT, securing semantic retention has not been sufficiently studied or implemented in state-of-the-art style transformation applications. We believe that this constitutes the most important challenge for the field going forward.
We further suggest that increased interaction between different fields would likely prove useful. While we have focused on style transformation from the perspective of information security, the field of \textit{automatic paraphrasing} is much broader in scope \cite{Madnani:Dorr2010, Lietal2018}, involving tasks such as automatic text simplification \cite{Siddharthan2010, Siddharthan2011, Wubbenetal2012, Narayan:Gardent2014}, controlling for style in MT \cite{Sennrichetal2016}, politeness transformation \cite{Rao:Tetreault2018}, or generating exercises for language pedagogy \cite{Baptista2016}. Systematically examining the effects of methods developed for other purposes on style obfuscation would constitute a valuable addition to the field.

\bibliographystyle{plain}
\bibliography{Stylometry_refs}

\end{document}